\documentclass[10pt,twocolumn]{article}

\usepackage{arxiv}
\usepackage[numbers,sort]{natbib}
\usepackage[utf8]{inputenc}
\usepackage[T1]{fontenc}
\usepackage{hyperref}
\usepackage{url}
\usepackage{booktabs}
\usepackage{amsmath,amssymb,amsthm}
\usepackage{mathtools}
\usepackage{bm}
\usepackage{graphicx}
\usepackage{dblfloatfix}  %
\usepackage{float}        %
\usepackage{subcaption}
\usepackage{xcolor}
\usepackage{microtype}
\usepackage{cleveref}
\usepackage{algorithm}
\usepackage{algorithmic}
\usepackage{tikz}
\usetikzlibrary{positioning,arrows.meta,fit,backgrounds,calc,shapes.geometric}

\graphicspath{{figures/}}

\definecolor{cVid}{RGB}{52,101,164}
\definecolor{cEnc}{RGB}{92,160,215}
\definecolor{cLat}{RGB}{200,143,20}
\definecolor{cLNN}{RGB}{180,60,30}
\definecolor{cDec}{RGB}{55,130,80}
\definecolor{cRen}{RGB}{120,70,150}
\definecolor{cForce}{RGB}{170,30,30}
\definecolor{cProj}{RGB}{150,85,10}

\newcommand{\xmark}{\ensuremath{\times}}
\newcommand{\pmark}{\ensuremath{\sim}}
\definecolor{cRkl}{RGB}{200,30,120}

\tikzset{
  bloc/.style={rounded corners=4pt, draw=#1, thick, fill=#1!15,
               minimum height=1.05cm, text centered, font=\small\bfseries,
               text=#1!60!black},
  arr/.style={-{Latex[length=2.2mm,width=1.6mm]}, thick, #1},
  lbl/.style={font=\scriptsize, #1},
  sub/.style={font=\tiny\itshape, text=black!50},
}

\newcommand{\R}{\mathbb{R}}
\newcommand{\q}{\mathbf{q}}
\newcommand{\dq}{\dot{\mathbf{q}}}
\newcommand{\ddq}{\ddot{\mathbf{q}}}
\newcommand{\f}{\mathbf{f}}

\newcommand{\Enc}{\operatorname{Enc}}
\newcommand{\Dec}{\operatorname{Dec}}

\DeclareMathOperator{\diag}{diag}

\title{LaGSplat: Inferring Physics-Governed Interactive Simulation\\
from Monocular Video Using Latent Lagrangian Gaussian Splatting}

\author{
  Louen Pottier\\
  Université Paris-Saclay, CEA, List, F-91120 Palaiseau, France\\
  \texttt{louen.pottier@cea.fr}
}

\date{\today}

\hypersetup{
  pdftitle={LaGSplat: Inferring Physics-Governed Interactive Simulation from
            Monocular Video Using Latent Lagrangian Gaussian Splatting},
  pdfauthor={Louen Pottier},
  pdfkeywords={Gaussian Splatting, Lagrangian neural networks, physics-based
               simulation, monocular video, interactive simulation},
}

\makeatletter
\let\orig@includegraphics\includegraphics
\renewcommand{\includegraphics}[2][]{%
  \IfFileExists{figures/#2}{\orig@includegraphics[#1]{#2}}{%
    \fbox{\begin{minipage}[c][2.4cm][c]{0.86\linewidth}%
      \centering\ttfamily\footnotesize [figure omitted in build:\\ \detokenize{#2}]%
    \end{minipage}}}}
\makeatother

\begin{document}

\twocolumn[{%
  \begin{@twocolumnfalse}
    \maketitle
    \thispagestyle{empty}
    \begin{abstract}
We present \textbf{LaGSplat} (Latent Lagrangian Gaussian Splatting), a
framework that infers interactive, physics-governed dynamics from one or a
few monocular videos. At inference it lets a user push on the filmed object,
rigid or deformable, with an external force that was never
measured, annotated, or seen during training. This is possible because a
low-dimensional latent state $\q \in \R^d$ plays two roles at once: it is
the generalised coordinate of a learned dissipative Lagrangian and the
conditioning variable of a Gaussian Splatting decoder. The inductive bias of this decoder, whose primitives are explicit
points $\mu_i(\q)$ that move with the object, is
what lets a force $f$ applied in the image pull back into a latent
generalised force $J(\q)^\top f$ and enter the equations of motion, which
pixel-space (CNN) or neural-field (NeRF) decoders cannot do. We validate LaGSplat on test cases of increasing difficulty,
from rigid to deformable and from autonomous to forced real systems,
combining monocular video and sensor measurements. We further demonstrate
interactive use: forces of arbitrary magnitude and direction can be applied to
the reconstructed object at any time, its response rendered in real time, in 2D
or 3D. Assuming a dissipative Euler-Lagrange equation over a few generalised
coordinates trades generality for a bounded, plausible response to unseen
forces, where an unconstrained predictor diverges.
    \end{abstract}
    \vspace{0.2em}
    \begin{center}
      \small Interactive demo:
      \url{https://louenpottier.github.io/lagsplat.html}
    \end{center}
    \vspace{0.3em}

    \begin{center}
      \includegraphics[width=\textwidth]{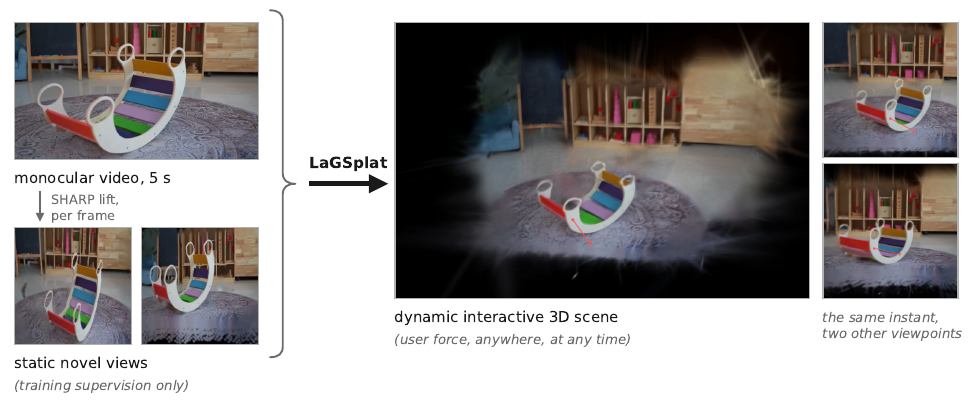}
      \captionof{figure}{%
        \textbf{From one monocular video to an interactive 3D scene.}
        Five seconds are filmed, from a single viewpoint;
        SHARP~\cite{mescheder2025sharp} renders novel views of every frame from
        other viewpoints. LaGSplat does not just replay that clip in 3D: the
        scene can be simulated from unseen initial conditions, and a force
        applied anywhere in it at any time (red arrow).}
      \label{fig:teaser}
    \end{center}
    \vspace{0.5em}
  \end{@twocolumnfalse}
}]

\section{Introduction}
\label{sec:intro}

This work sits at the intersection of two active research areas that appear,
at first glance, unrelated. On one side, dynamic scene reconstruction from
video has reached real-time photorealism
\cite{kerbl20233dgs,yang2024realtimephoto4dgs,wu20244dgs}, but these
representations are parameterized by time $t$; they replay the recorded clip
and no interaction with the objects in the scene is possible. On the other
side, learning to simulate physical systems from data
\cite{cranmer2020lnn,greydanus2019hnn,lutter2019delan} recovers structured
equations of motion that can be forced, extrapolated, and controlled, but
training requires a dataset recording the time evolution of the system's
generalized coordinates, observed through sensors or produced by a simulator.
We aim for a single framework unifying both capabilities: from video alone,
rather than from simulation or sensor data, a trainable architecture that
recovers the governing equations of the filmed system (such as a pendulum, a
deformable structure, or a soft robot) and reconstructs the scene
photorealistically, in a way that allows interaction at inference, changing
physical coefficients (damping, inertia) or applying a force anywhere on the
system at any time, and letting it evolve in a physically plausible way.

A large and fast-growing body of work already generates videos of systems
that obey physical laws: by attaching a prescribed physical engine to
a reconstructed or diffusion-generated scene
\cite{xie2024physgaussian,zhang2024physdreamer,liu2024physgen,yuan2026newtongen},
by learning a world model that generalizes across many systems
\cite{geng2025neurok,xiao2026lawm}, or by conditioning a video generator
directly on physical control signals \cite{gillman2025forceprompting}. These methods synthesize plausible
dynamics for a new scene, often from as little as a single image or a static
shape, but the motion is never registered against the measured behavior of
the particular system depicted. Much rarer is the opposite setting, the one
we adopt, in which the physics of a single filmed system is identified
from its own video, a counterpart on pixels of model reduction and of
learning equations of motion from simulated states
\cite{sharma2024lopinf,cranmer2020lnn,agrawal2026llnn}. Only a handful of
works reach it from video
\cite{greydanus2019hnn,krauss2026von,castaneda2025learning,quan2025particlegs},
and each gives up one of the ingredients we need for what is the central
contribution of this work: applying, at inference, a force at an arbitrary
point of the scene and letting the identified dynamics carry it through the
whole system.

Except from video, such a forcing mechanism is not new in itself: wherever a
dynamics is integrated in a low-dimensional latent space, a
force applied in the full configuration space transports into a latent
generalized force through the decoder (or encoder) Jacobian and enters the
right-hand side of the equation of motion
\cite{fulton2019latentspace,agrawal2026llnn,friedl2025riemannian,friedl2026hamscale}.
Carrying the mechanism over to video asks for two ingredients at once:
an Euler-Lagrange equation learned over a low-dimensional latent space, and
an image decoder made of \emph{material points} that move with the object, so that the Jacobian of
those points is what pulls an applied force back into the latent space. No
method that identifies physics from video offers both. 
The learned dynamics is either an unconstrained neural ODE with no
mechanical structure \cite{quan2025particlegs}, or fixed in
advance to a prescribed form rather than a general equation of motion
\cite{krauss2026von,castaneda2025learning,wang2026contactgaussian}. And where a genuine law is within
reach, the explicit material point is the missing ingredient instead: the scene is carried on
fixed pixels \cite{greydanus2019hnn,krauss2026von} reconstructed frame by
frame, with no primitive that persists as an explicit point moving with the
object, or it is not reconstructed at all \cite{castaneda2025learning}. No single work both
identifies a genuine mechanical law and renders it over explicit,
moving primitives. LaGSplat fills this gap.

The key to LaGSplat is that a single low-dimensional latent state
$\q\in\R^d$ plays two roles at once: it is
the generalized coordinate of a learned dissipative Lagrangian, and it is
the conditioning variable of a Gaussian Splatting decoder whose primitives
are explicit points $\mu_i(\q)$ that move with the object
(Fig.~\ref{fig:overview}).
An encoder maps raw frames to $\q$, a latent
Lagrangian neural network integrates the learned equations of motion, and
the decoder renders each $\q$ back into an image, in 2D or, under depth or
multi-view supervision, from novel 3D viewpoints. Nothing supervises $\q$
itself: it emerges from the photometric loss and the Euler-Lagrange
residual acting on the same encoder, with no annotation of the state and no
external simulator. The mechanism is the one that discovers the latent state of a
reduced-order Lagrangian or Hamiltonian network
\cite{agrawal2026llnn,friedl2025riemannian,friedl2026hamscale}, with the
reconstruction loss now photometric, on pixels rather than on generalized
coordinates.

Because the decoder's primitives are explicit points, a variation of the
latent state moves them by $\delta\mu = J(\q)\,\delta\q$ with
$J=\partial\mu/\partial\q$, and this is what
makes the reconstructed object pushable: by virtual work, a force $f$
applied on the object in the image transports into a latent generalized
force $J(\q)^\top f$ that enters the right-hand side of the same equations
of motion without retraining. That a model trained on video in which no
force was ever applied can respond to one is counter-intuitive, but it follows
from what the Jacobian is: a
property of the
kinematics $\q\mapsto\mu(\q)$ alone, not of the loading, so the autonomous
video that trains the decoder already determines it. What autonomous motion
leaves undetermined is a single scalar, the physical unit of the force to be applied
in the original space. The learned Lagrangian and its dissipation can be
rescaled together without changing any observed trajectory, and that same
factor sets that unit. Provided the dynamics is well identified from the video
and the primitives move correctly with the object, the learned response to an
arbitrary force, as long as it keeps the state $\q$ within the range visited
in training, is therefore correct in
direction and in shape, and determined in amplitude only up to that single
global factor. A measurement relating an applied force to the resulting
displacement fixes it once for the whole model, and absent such a measurement
it remains a free overall scale, chosen by hand. One further boundary is
structural rather than a matter of scale: only the motions the video actually
exhibits enter $\q$ in the first place. A part of the object that stays fixed
throughout the clip, or a deformation mode the video never excites, corresponds
to a latent direction the training data never populates; the model carries no
coordinate for it, so at inference it behaves as infinitely rigid, and any
force applied there leaves the scene unchanged.

LaGSplat sits at the intersection announced above: from dynamic scene
reconstruction it inherits the explicit Gaussian primitives and their
real-time differentiable rasterizer, and from learned analytical mechanics the
latent Lagrangian and the generalized force that drives it. Our contribution
is this pipeline itself: a fully differentiable path from a
monocular video to an identified physical model and a real-time interactive
simulation of the filmed object, rendered in 2D or 3D and responsive to
forces never measured. The Lagrangian prior narrows the scope to systems with a
small number of generalized coordinates, which is
precisely the regime where it pays off: we show that a motion predictor
without a mechanical prior \cite{zhao2024gaussianprediction} fails to reach a
stable equilibrium that the training video does not contain, whereas LaGSplat
converges to one by construction. We validate the pipeline on cases of increasing
difficulty, from rigid to deformable and from autonomous to forced:
a simple pendulum, an oscillating rigid body, a hanging bag, and the
pneumatic actuator of \citet{krauss2026von}. The scene is reconstructed in 2D,
or in 3D under multi-view supervision, and responds to forces in real time in
both settings.

\section{Related Work}
\label{sec:related}

\begin{table*}[!t]
\centering
\caption{%
  \textbf{Positioning on the five axes.} \checkmark: the axis is met,
  \pmark: partly met, \xmark: not met, ---: not applicable.}
\label{tab:positioning}
\setlength{\tabcolsep}{4pt}
\small
\begin{tabular*}{\textwidth}{@{\extracolsep{\fill}}lccccc@{}}
\toprule
Method & \shortstack{Physics-governed\\dynamics} & \shortstack{Learned\\latent state}
  & \shortstack{Own-video\\supervision} & \shortstack{Material-point\\kinematics $\mu_i(\q)$}
  & \shortstack{Interactive force\\at inference} \\
\midrule
\multicolumn{6}{l}{\emph{Dynamic novel-view synthesis (\S\ref{sec:related-rendering})}}\\
\quad Dynamic NeRF \citep{pumarola2021dnerf,park2021nerfies}
  & \xmark & \xmark & \checkmark & \xmark & \xmark \\
\quad 4DGS family \citep{yang2024realtimephoto4dgs,wu20244dgs,wang2025freetimegs}
  & \xmark & \xmark & \checkmark & \checkmark & \xmark \\
\midrule
\multicolumn{6}{l}{\emph{Learned analytical mechanics (\S\ref{sec:related-mechanics})}}\\
\quad Neural ODE \citep{chen2018neuralode}
  & \xmark & \xmark & \xmark & --- & \xmark \\
\quad Latent neural ODE \citep{iakovlev2023shooting}
  & \xmark & \checkmark & \xmark & \xmark & \xmark \\
\quad PINN \citep{raissi2019pinn}
  & \pmark & \xmark & \xmark & \xmark & \xmark \\
\quad LNN / HNN / DeLaN \citep{cranmer2020lnn}
  & \checkmark & \xmark & \xmark & --- & \pmark \\
\quad Structure-preserving ROM \citep{sharma2024lopinf}
  & \checkmark & \pmark & \xmark & \pmark & \checkmark \\
\quad Latent Lagr.\ / Ham.\ nets \citep{agrawal2026llnn,friedl2025riemannian,friedl2026hamscale}
  & \checkmark & \checkmark & \xmark & \pmark & \checkmark \\
\midrule
\multicolumn{6}{l}{\emph{Generating physically plausible video (\S\ref{sec:related-generative})}}\\
\quad NewtonGen \citep{yuan2026newtongen}
  & \pmark & \xmark & \pmark & \xmark & \xmark \\
\quad LaWM \citep{xiao2026lawm}
  & \checkmark & \checkmark & \pmark & \xmark & \xmark \\
\quad GS + MPM \citep{xie2024physgaussian}
  & \pmark & \xmark & \xmark & \checkmark & \pmark \\
\quad PhysDreamer \citep{zhang2024physdreamer}
  & \pmark & \xmark & \pmark & \checkmark & \pmark \\
\quad PhysGen \citep{liu2024physgen}
  & \pmark & \xmark & \pmark & \xmark & \pmark \\
\quad Force Prompting \citep{gillman2025forceprompting}
  & \xmark & \xmark & \pmark & \xmark & \pmark \\
\quad VR-GS \citep{jiang2024vrgs}
  & \pmark & \xmark & \xmark & \checkmark & \checkmark \\
\quad NeuROK \citep{geng2025neurok}
  & \checkmark & \checkmark & \xmark & \checkmark & \checkmark \\
\midrule
\multicolumn{6}{l}{\emph{Learning and simulating from video supervision (\S\ref{sec:related-identify})}}\\
\quad Neural state variables \citep{chen2022statevariables}
  & \xmark & \pmark & \checkmark & \xmark & \xmark \\
\quad CpAE \citep{zhu2025cpae}
  & \xmark & \checkmark & \checkmark & \xmark & \xmark \\
\quad Pixel-HNN \citep{greydanus2019hnn}
  & \checkmark & \checkmark & \checkmark & \xmark & \xmark \\
\quad Casta\~neda et al.\ \citep{castaneda2025learning}
  & \pmark & \checkmark & \checkmark & \xmark & \xmark \\
\quad EvoGS / GaussianPred.\ \citep{asiimwe2025evogs,zhao2024gaussianprediction}
  & \xmark & \pmark & \checkmark & \checkmark & \xmark \\
\quad NVFi \citep{li2023nvfi}
  & \pmark & \xmark & \checkmark & \checkmark & \xmark \\
\quad Poking Plants \citep{blattmann2021ii2v}
  & \xmark & \pmark & \checkmark & \xmark & \pmark \\
\quad ParticleGS \citep{quan2025particlegs}
  & \xmark & \checkmark & \checkmark & \checkmark & \xmark \\
\quad Krauss VON \citep{krauss2026von}
  & \checkmark & \checkmark & \checkmark & \xmark & \pmark \\
\quad ContactGaussian-WM \citep{wang2026contactgaussian}
  & \pmark & \pmark & \pmark & \checkmark & \checkmark \\
\midrule
\textbf{LaGSplat (ours)}
  & \checkmark & \checkmark & \checkmark & \checkmark & \checkmark \\
\bottomrule
\end{tabular*}

\vspace{3pt}
\begin{minipage}{\textwidth}
{\footnotesize
  \emph{Physics-governed dynamics}: the
  state evolves under a mechanical law (Lagrangian or Hamiltonian), so
  the model extrapolates in time and responds to forces rather than
  merely interpolating the observed motion; \checkmark: a generic mechanical
  class (an Euler-Lagrange or Hamiltonian law) whose content is
  learned from data; \pmark: the governing law of the specific
  system is supplied a priori, at most calibrated, or enters only as a
  soft residual; \xmark: no mechanical structure (time indexing or a
  free ODE). \emph{Learned latent state}: the dynamics evolve on a
  compact state discovered from data. \checkmark: a low-dimensional
  $\q$ is learned; \pmark: a reduced state is used but partly imposed
  (known coordinates, tracked keypoints, or a fixed linear reduction), or
  discovered from data yet not what the dynamics is stepped over;
  \xmark: no learned latent (raw-space or time-indexed dynamics).
  \emph{Own-video supervision}: the model is trained on the video of the very
  system it then simulates.
  \checkmark: monocular video of that system alone; \pmark: its video plus an
  auxiliary signal, stand-in frames in place of its own footage (synthetic
  renders, diffusion-generated frames, or side states), or a prior learned
  across the videos of other systems; \xmark: no video supervision at all (the
  model learns from simulated trajectories, measured states, or meshes
  instead).
  \emph{Renders through material points}: the renderer reads explicit
  primitives $\mu_i(\q)$ that move with the object. \checkmark: the image is
  rasterized from such primitives; \pmark: the decoder does output material
  points (mesh nodes, lumped masses) but no image is rendered through them;
  \xmark: an Eulerian decoder, a field or a warp
  evaluated on fixed pixels; ---: no spatial decoder at all, the model never
  leaving state space. \emph{Interactive
  force at inference}: a new external force, absent from training, enters the
  dynamics at inference and propagates through the whole system, without
  retraining. \checkmark: applicable anywhere on the object; \pmark:
  restricted, to a few predefined points, to the latent state alone, or to the
  initial condition; \xmark: no force enters.\par}
\end{minipage}
\end{table*}

\enlargethispage{\baselineskip}

Four families frame the discussion. Dynamic novel-view synthesis has the
images but no equation of motion (\S\ref{sec:related-rendering}). Learned
analytical mechanics has the equations but no image
(\S\ref{sec:related-mechanics}). The last two put the two together, and
differ in what the video is for them, an output or a supervision: generating
physically plausible video renders motion for a scene it was never trained to
reproduce (\S\ref{sec:related-generative}), while learning and simulating
from video supervision fits a dynamics to the footage of the very system
depicted (\S\ref{sec:related-identify}). The last setting is the one LaGSplat
adopts.

Five design choices, laid out in Section~\ref{sec:intro}, place each family
on the grid of Table~\ref{tab:positioning}. Does the dynamics carry a
structured mechanical prior, or is it a free ODE? Is the state an observed
physical coordinate, or a learned latent state? Is the model supervised by the
video of the very system it simulates, or by a substitute, a prior learned
across other systems, simulated trajectories, or measured states? Does
rendering go through explicit points that move with the object, or through a
field painted on fixed pixels? Can an external force be injected into the
dynamics at inference? The third question is what separates the last two
families: only one of them is supervised by the system's own video, which is
why generated footage and cross-system priors score \pmark\ there. Each family below
holds some of these cells and misses others.

\subsection{Dynamic novel-view synthesis}
\label{sec:related-rendering}

Our decoder architecture comes from this family. Its aim is to synthesize
views of a moving scene indexed by the time $t$, where ours are indexed by
the state $\q$; the goal differs, but the design space of the decoder is the
same one, and we select in it on a criterion of our own. Applying an effort
to the scene at inference (Section~\ref{sec:force}) requires a decoder whose
primitives are particles followed in their motion, each carrying its own
position $\mu_i(\q)$, since a force has a point of application and needs
somewhere explicit to attach.

That criterion rules out the dominant form of the radiance-field branch.
NeRF-based methods \cite{mildenhall2020nerf} animate the field by
conditioning it on time or by deforming a canonical volume
\cite{pumarola2021dnerf,park2021nerfies}: they answer what colour lies at a
given point of the image, following no piece of the object from one frame to
the next. Such a decoder is \emph{Eulerian}, sampling a field over a fixed
support rather than transporting matter through it \cite{pfaff2021meshgraphnets},
and a force has nothing to attach to (the same obstruction returns, on fixed
pixels, in Section~\ref{sec:related-identify}). The branch is not uniformly
so: a deformation field written from the object outwards
\cite{guo2023forwardflow} and a particle-based radiance field
\cite{lin2025dapnerf} do carry material points, and PAC-NeRF
\cite{li2023pacnerf}, whose MPM advection needs moving particles while its
rendering needs a fixed field, carries both at once at the price of
particle-to-grid transfers at every step. Their decoders would have been
usable for our purpose. We took the other branch, which supplies the same
material points and forms the image by rasterizing them rather than by
evaluating a network at many samples along every ray. Since the simulation is
driven interactively, the applied force changing the state between one frame
and the next, real-time rendering is a requirement of the setting and not a
convenience.

In 3D Gaussian Splatting \cite{kerbl20233dgs} a scene is a set of explicit
anisotropic Gaussian primitives, rasterized differentiably in real time, and
one and the same object is the material point and the rendered primitive.
Among its dynamic extensions we build on the higher-dimensional Gaussian of
\citet{yang2024realtimephoto4dgs}, whose extra axis is conditioned away at
render time by a Schur complement, and we rasterize with \texttt{gsplat}
\cite{ye2024gsplat}. The alternative route, a neural network warping
canonical primitives \cite{wu20244dgs}, meets the criterion just as well: it
too gives every primitive a position, hence a Jacobian, obtained by
differentiating the network. We set it aside on two counts. It is not the
better posed of the two on motion-rich content, where FreeTimeGS
\cite{wang2025freetimegs} finds primitives placed directly in the extended
space preferable; and extending the primitive itself keeps the map $\q \mapsto
\mu_i$ affine, so that the Jacobian $J_i = \partial\mu_i/\partial\q$ of each
Gaussian is a constant matrix stored alongside it. An interactive effort then
enters in closed form, rather than through automatic differentiation across
the decoder at every time step. The first count holds on our own data too, in a
comparison at equal budget reported in Appendix~\ref{sec:app-deform}.

Our one change is to what that extra axis carries. Where these methods put the
time, we put the $d$ latent coordinates, so the block conditioned away by the
Schur complement is a whole latent block and the ellipsoid actually rasterized
is the one conditioned on the current state $\q$. Gaussians have been lifted to
higher dimensions before, for fitting capacity alone, with no dynamical meaning
attached to the added axes \cite{diolatzis2024ndgaussians,serifi2026hypergaussians};
what is new here is not the dimension but that those axes carry a state ruled by
an equation of motion. Nothing in this family says how $\q$ evolves, and the
next one supplies that.

\subsection{Learned analytical mechanics}
\label{sec:related-mechanics}

Our dynamics comes from this family, as our decoder came from the last one.
Its members learn their physics from measured or simulated states, where we
have to learn ours from images. That change of data does not bear on the
mechanics; it bears on where the state and the force come from, both of which
reach us through the decoder rather than from a simulator. We ask three things
of a candidate: that the equation of motion be the learned object itself, that
it be written over a state discovered from the data, and that it expose a
right-hand side where an external force can be added at inference.

On the first, the family spans two poles. Neural ODEs \cite{chen2018neuralode}
fit the transition rule in full generality, the \xmark\ anchor of the prior
axis of Table~\ref{tab:positioning}, and learned simulators sit at that same
free end, whether they advect mesh nodes or particles
\cite{battaglia2016interaction,li2019dpinet,pfaff2021meshgraphnets,sanchezgonzalez2020gns}
or read a field over a fixed domain
\cite{kim2019deepfluids,meyer2021deepsurrogate,li2021fno}. PINNs
\cite{raissi2019pinn} anchor the supplied side, the residual of a known
PDE regularizing a field network, which is what the \pmark\ of that axis
denotes. Between them, Lagrangian and Hamiltonian neural networks learn the law
itself, building conservation and dissipation into it
\cite{cranmer2020lnn,greydanus2019hnn,lutter2019delan,finzi2020simplifying,bhattoo2022lagrangian},
which meets the first requirement. They read their coordinates from joint
encoders or simulated states, leaving the second to the branch below.

A recent branch writes the same machinery over a learned latent state
\cite{pottier2024lebnn,agrawal2026llnn,friedl2025riemannian,friedl2026hamscale},
or over a linear POD reduction \cite{sharma2024lopinf}. What is new there is
the learned law, not the latent integration: graphics got to the latter first
with the law left prescribed, \citet{fulton2019latentspace} time-stepping the
true elastodynamic equations in an autoencoder subspace at interactive
rates on a 164k-element mesh, with the stiffness pulled back as $\tilde H =
J^\top \tilde K_0 J$ (its descendant \citep{li2025selfsupervised} learns the
integrator instead, and names captured data and Gaussian-splat
reconstructions as its natural next step). These models also admit a
generalized force. L-LNN \cite{agrawal2026llnn} transports supervised forces
into the latent space through an encoder Jacobian, and the reduced-order models
of Friedl et al.\ \citep{friedl2025riemannian,friedl2026hamscale} pull the
full-space force back through the decoder Jacobian, the same virtual-work
transport we use. All three requirements are thus met here, but on data we do
not have: those forces are measured in the full configuration space of a
simulator, and the state is instrumented in the same way, read from joint
encoders, simulator states, or a linear reduction of a known mesh. Ours have to
come through the decoder, from pixels. The transport is inherited, its
provenance is not, and the identifiability that instrumentation buys is left on
our side.

What we take from here is the structure, a learned law over a learned state,
together with the $J^\top$ pull-back that the reduced-order members already
use, meaningful in them for the reason it is in us, because their decoders too
land on material points. Learning such a model beyond one degree of freedom is
known to be delicate, and recent work isolates the failure modes we also meet:
the ill-posedness of identifying a structured model when only positions are
observed and the momenta stay latent \cite{bhardwaj2026phast}, non-convex
potentials \cite{jones2025multiwell}, ill-conditioned mass matrices
\mbox{\cite{hamzaogullari2026stabilizedlnn,kotecha2025lnnquadruped}}, and the
insufficiency of single-step derivative matching for long-horizon stability
\cite{iakovlev2023shooting,laiche2025noaccel,hansen2025dflnn}; the training
recipe of Section~\ref{sec:training} builds on these lessons. What no member of
this family does is read those material points off a video, or render an image
through them. Images arrive with the next two families.

\subsection{Generating physically plausible video}
\label{sec:related-generative}

This family assembles the same two components we do, a spatial decoder
(\S\ref{sec:related-rendering}) driven by an equation of motion
(\S\ref{sec:related-mechanics}), but its dynamics is brought to the scene
rather than learned from its video. It only has to look right, not to agree
with the recorded behavior of a particular object, and the family splits into
two branches on where that dynamics comes from.

One branch prescribes the law. PhysGaussian and its successors delegate the dynamics
to an external MPM simulator with a predefined material prior
\cite{xie2024physgaussian,lin2025omniphysgs,le2025pixie,jiang2016mpm}. The
momentum equation carries an external-force slot and the particles are
material points, but in most of the family the demonstrated interventions are
scripted initial velocities on hand-set materials, so that slot stays
unexercised though structurally available.
VR-GS \cite{jiang2024vrgs} exercises that slot in
real time, a user in virtual reality grabbing, dragging and colliding the object
at interactive rates, so the effort is sustained along the trajectory instead of
being posed as an initial condition. Its material is set by hand rather than
fitted, and the capture it starts from is a static multi-view scan, so nothing
of the dynamics is read from a recording of the object in motion.
PhysDreamer
\cite{zhang2024physdreamer} goes furthest on the data side. It calibrates a stiffness field
against a clip generated by a video-diffusion prior, then answers pokes and
drags inside the MPM simulator, without retraining, on a material fitted to
data rather than set by hand. That is our nearest neighbor on that side, and
the response it produces is a genuinely coupled elastic one. But the
intervention is still posed as an initial condition, an impulse over a region
designated on the object, followed by an autonomous rollout, not a force
sustained along the trajectory. It differs elsewhere too: the law stays
prescribed (only the stiffness is calibrated), no low-dimensional state is
learned, the supervising video is generated rather than observed, and the
response costs about a minute of simulation per rendered second, where a
latent Lagrangian integrates $d$ scalar equations in real time.

PhysGen rolls out a prescribed 2D rigid-body engine, NewtonGen learns the
coefficients and a nonlinear residual of a fixed second-order Newtonian
skeleton, and both render through a video-diffusion model
\cite{liu2024physgen,yuan2026newtongen}. Neither is forced along the
trajectory. PhysGen takes a user force as its headline input, but it enters
the 2D rigid engine as a single initial impulse at the center of mass, before
the rollout; NewtonGen's control is declarative only, initial conditions
parsed from the prompt into an autonomous ODE that admits no external force
at all.

Force Prompting \cite{gillman2025forceprompting} goes one step further and
keeps no law at all: it answers localized pokes and global wind fields and
generalizes to unseen geometries, with no mechanical structure anywhere. Its
response is a conditioning channel trained on force-video pairs, so every kind
of effort has to be exhibited at training time; and the effort itself is one
vector, a magnitude and an angle, fixed before the clip is generated and
unrolled over its frames. The model is neither
real-time nor per-frame causal. No force can be redefined while the motion
unfolds, which is what holds it to the same partial mark as the rest of this
branch. The mechanical structure they do without is what spares us both
restrictions: we never observe a force at all, and ours is an arbitrary
$f(t)$, redefinable at any instant, deduced from the identified law and carried
by virtual work through the material points into the right-hand side that
$J^\top f$ shares with the actuation term of Section~\ref{sec:pressure}. That transport is
what a material point buys: not the ability to answer a force, which a control
field distributed over a fixed grid has too \cite{holl2020controlpdes}, but the
ability to answer, interactively, one that was never seen.

A second branch learns a transferable prior instead, and is judged on what
carries over to systems it has never seen rather than on its agreement with a
particular recorded object. LaWM
\cite{xiao2026lawm} learns a discrete latent Lagrangian whose variational
integrator is the transition rule, symplectic and drift-free over long
horizons. But it trains on simulator renders, keeping simulator states or
robot signals as auxiliary supervision when available; its learned Lagrangian
is conservative, and its rollout unforced, since control inputs enter its
embodied benchmark as conditioning signals only, not as a generalized force.
At the unstructured end of that same aim, JEPA-style world models drop the
mechanics altogether and forecast embeddings for planning, adapting online at
test time \cite{wang2026adajepa}.

Closest to us in this branch is NeuROK \cite{geng2025neurok}. Building on CANOR
\cite{he2025canor}, it learns a latent kinematic parameterization of meshes
and integrates a latent Euler-Lagrange equation whose kinetic metric is the
pullback of the decoder Jacobian, $G(\q)=J^\top J$, the same geometric
construction our force transport is built from. The shared object is that
pullback, not the decoder: NeuROK renders a deforming mesh, whose material
points are its vertices under a globally nonlinear map, not the affine
Gaussian primitives of our splatting branch. The kinship extends past the
metric to the force: its supplementary material forces that equation through
that same pullback, $Q = J^\top F$ over a region and a time window of the
user's choosing, and reports the response against force magnitude, which makes
NeuROK the one member of this family to hold the force column outright. What
separates us is not the mechanism but the data: NeuROK is not supervised from
video but learned across large 4D datasets of deforming meshes, curated and
simulated, a learned prior meant to transfer to new shapes rather than a
mechanics identified from one system's own footage. It is thus a structural
cousin on the far side of the video-supervision axis, not a direct from-video
competitor.

Across this family the physics does reach images, and one member, NeuROK,
forces its equation through the same virtual-work transport we use. But the
governing law is prescribed or transferred, never identified from the video of
the very system to be simulated, as it is in the family that follows.

\subsection{Learning and simulating from video supervision}
\label{sec:related-identify}

Change the data and \S\ref{sec:related-mechanics} becomes this last family. Its
members share the aim of learned analytical mechanics, an equation of motion
learned over a state discovered from the data, one system at a time, but what
they are given is video of that system rather than measured or simulated
states. From it they identify and simulate the behavior of the particular
object filmed, not a plausible behavior for a scene of that kind, which is the
break from \S\ref{sec:related-generative}. What they gain over
\S\ref{sec:related-rendering} is that the recording stops being the only thing
that can be played back: a state and a law let the motion be restarted from an
initial condition that was never filmed, and continued past the last frame of
the clip. This is the setting LaGSplat adopts, and the works that share it
build both halves from pixels alone, a decoder from the design space of
\S\ref{sec:related-rendering} and a dynamics from that of
\S\ref{sec:related-mechanics}. 

The first branch learns an evolution rule with no mechanical structure in it, a
transition map, a velocity field or a free ODE, never a Lagrangian or a
Hamiltonian. The lineage opens with \citet{chen2022statevariables}, who recover from video how many
state variables a filmed system has, and one admissible set of them: a network
trained to predict the next pair of frames, an intrinsic-dimension estimate on
its bottleneck~\citep{levina2004intrinsic}, rounded to the nearest even integer
since positions and velocities come in pairs, and a second autoencoder
compressing that bottleneck to the resulting width. Those variables are never
stepped: the prediction runs through the pixels, regenerating a frame and
re-encoding it at each step, and the compact state only projects it back onto
the manifold to keep it from drifting. \citet{zhu2025cpae} remove what makes such a state
unusable by a continuous law. A displacement smaller than one pixel leaves the
image almost unchanged, so a latent read off pixels tends to advance in jumps
where the motion is smooth. Their remedy is a Lipschitz condition on the filters of the
first layers, allowing an ODE to then be fitted directly in the latent space,
the encoder frozen, where the practice is to optimize the latent law jointly
with the maps that produce its coordinates
\cite{greydanus2019hnn,quan2025particlegs,krauss2026von}, the dynamics
objective bending the latent towards coordinates its model class can fit. The
dynamics they plug in is a plain neural ODE, the same choice ParticleGS
\cite{quan2025particlegs} makes over a handful of scene-wide motion modes,
shared by every Gaussian, its rule unconstrained and carrying no physical
parameter. The remaining members change the form of that rule rather than its
nature. EvoGS \cite{asiimwe2025evogs} integrates a learned
velocity field that still reads the time explicitly. GaussianPrediction
\cite{zhao2024gaussianprediction} forecasts key-point trajectories with a
graph network. NVFi \cite{li2023nvfi} constrains a dense velocity field with
a PDE residual, a supplied prior in the sense of \S\ref{sec:related-mechanics},
but one written on a kinematic field rather than on an inertia.
\citet{blattmann2021ii2v} learn from real videos how an object deforms after a poke at one pixel, a displacement to
reach rather than a force. None of these
rules carries an energy, an inertia or a dissipation, so none exposes a
right-hand side into which an external effort could enter, and nothing keeps
what they extrapolate past the training video physically plausible, as
GaussianPrediction illustrates (Section~\ref{sec:exp-extrapolation}).

The rest do write a mechanical law, and differ by
how much of it is left to be learned. The pixel variant of HNN \cite{greydanus2019hnn} learns a
latent state from synthetic pendulum images and its Hamiltonian jointly with
the autoencoder: the conservative pole of
\S\ref{sec:related-mechanics} over a learned latent state, one coordinate and
its momentum for the pendulum it is demonstrated on. It is our scheme
without the two things we most need from it, since a conservative and unforced
Hamiltonian receives no force in its right-hand side, and a fixed-pixel
decoder exposes no material point for one to act on.
\citet{castaneda2025learning} sit at the supplied end of that same axis,
estimating the scalar coefficients of a known governing equation, a
linear second-order ODE, from monocular video. Their footage is real, which
pixel-HNN's is not, but the systems it shows have one degree of freedom each,
a pendulum angle, a decaying LED intensity, a falling height; and they take no
decoder at all, recovering the dynamics alone and reconstructing nothing.
Krauss et al.\ \cite{krauss2026von} carry the setting to systems with several,
learning a chain of latent oscillators from monocular video of real pneumatic
soft continuum robots, with measured chamber pressures as a known input. Their
dynamics is constrained to a linear model, and the only force it carries is
that pressure, a generalized force on a fixed actuation channel. Their decoder
is Eulerian in exactly the sense of \S\ref{sec:related-rendering}: it paints
intensities on a fixed pixel grid through position-wise operations, and the
per-oscillator attention centroids it reads out are a derived statistic, not a
generative material point. Their transpose is computable and the authors do
differentiate it, but it is the gradient of a heuristic centroid, which can
move because the attention field deforms rather than because matter translates,
and which nothing supervises to follow the same material. The model uses it
only to visualize forces, never to apply one, and applying one would
reach those centroids alone, half as many as the latent state has coordinates,
where an explicit decoder offers every primitive, hence any point of the
object.

LaGSplat is assembled from parts of these families. Its decoder is the
higher-dimensional Gaussian of \citet{yang2024realtimephoto4dgs}, rasterized
with \texttt{gsplat}, whose added axes we fill with the latent state instead of
the time and whose affine map $\q \mapsto \mu_i$ keeps every Jacobian $J_i$ a
constant matrix. Its dynamics is a dissipative Lagrangian over a learned latent
state, forced through the same virtual-work pull-back $J^\top$ that L-LNN
\cite{agrawal2026llnn} and Friedl et al.\
\cite{friedl2025riemannian,friedl2026hamscale} already apply, here on a Jacobian
read off the image decoder rather than off a simulator. Its encoder takes
from \citet{zhu2025cpae} the Lipschitz condition on the filters that makes the
latent trajectory continuous. The chart is fixed first on a reconstruction criterion,
the structured law is inferred on it afterwards. Letting photometric loss alone choose
the chart leaves it a scaling and a curvature the mechanics must absorb, which
is why our inertia is a state-dependent $M(\q)$ where pixel-HNN,
\citet{castaneda2025learning} and Krauss et al.\ carry a constant mass and let
the dynamics shape the chart instead; that nonlinearity earns its place on their
own pneumatic soft continuum robots, the public dataset we report on, where four
latent coordinates outperform their ten-dimensional oscillator chain on the
two-segment robot (Section~\ref{sec:experiments}). What is new is the pair. The
axes added to the primitive carry the very state the equation of motion rules,
so the points a force acts on and the coordinates that force integrates are the
same object.

An Eulerian decoder exposes no material point for a force to act on, and a free
dynamics leaves no right-hand side for it to enter. Krauss et al.\
are the informative case, the only ones to close one of the two without the
other, their right-hand side receiving a force their decoder gives nowhere to
be applied. \citet{wang2026contactgaussian} close both, but limit themselves to
rigid bodies where we also address deformable objects. Their Gaussian scene is
carried by prescribed equations of motion that do take an applied force at
inference. Only their coefficients and the collision geometry are fitted, and
the state stepped is the pose of bodies given in advance rather than one
discovered from the images.
The remaining cell is ours, a dissipative Lagrangian over a learned
latent state, supervised by video alone, rendered through explicit points and
driven at inference by an interactive force, identified for one filmed system
at a time: a deliberate scope, trading zero-shot transfer to unseen objects for
a model grounded in the observed system alone.

\section{Method}
\label{sec:method}

\subsection{Problem Statement and Overview}
\label{sec:problem}

\begin{figure*}[t]
\centering
\includegraphics[width=0.80\textwidth]{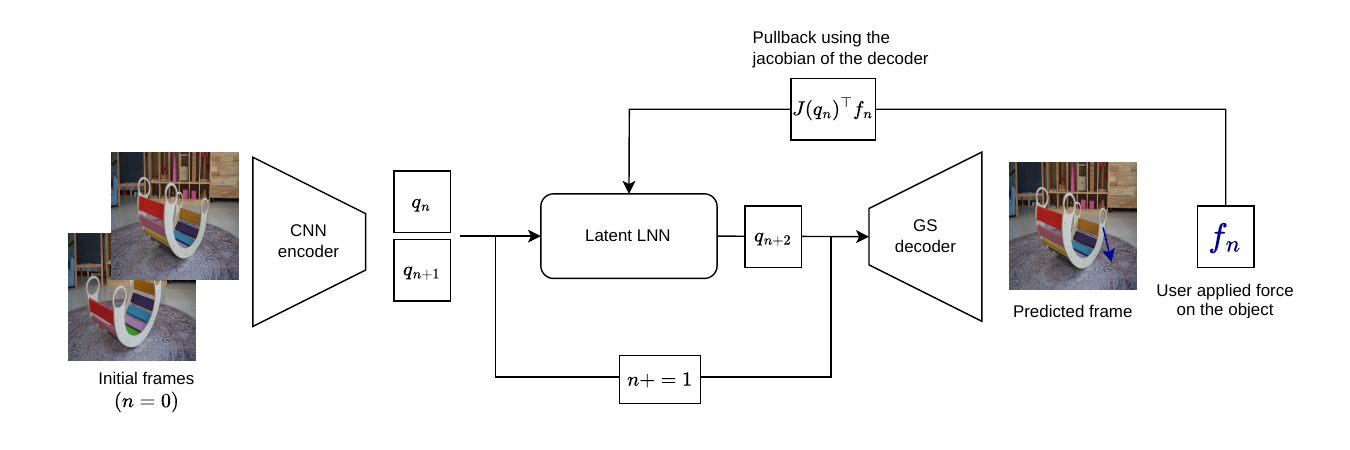}
\caption{%
  \textbf{Interactive simulation of a physical scene with a LaGSplat trained
  from video.}
  A CNN encoder maps a pair of initial frames to latent states
  $\q_0, \q_1$. The governing equation learned by a Lagrangian Neural
  Network (LNN) in the latent space is integrated in time. At any time a
  Gaussian Splatting decoder reconstructs the image from the latent state. A
  user force $f_n$ can be applied anywhere at any time on the image; it is
  pulled back through the decoder Jacobian into a latent generalized force
  $J(\q_n)^\top f_n$ that drives the LNN.}
\label{fig:overview}
\end{figure*}

Let $\{I_t\}_{t=1}^{T}$ be a monocular RGB video of a physical system with $d$
degrees of freedom, sampled at time step $\Delta t$. LaGSplat learns three
differentiable networks: an encoder $\Enc : \R^{C\times H\times W}\!\to\R^d$, a
latent Lagrangian $\mathcal{L} : \R^{2d}\!\to\R$ with dissipation
$\mathcal{D} : \R^{2d}\!\to\R$, and a Gaussian Splatting decoder
$\Dec : \R^d\times\mathrm{SE}(3)\to\R^{C\times H\times W}$, rasterizer included,
whose viewpoint argument is free only under multi-view supervision
(Section~\ref{sec:multiview}). The latent state $\q_t=\Enc(I_t)\in\R^d$, which
nothing supervises, is at once the generalized coordinate of the equation of
motion and the variable the decoder renders from. Figure~\ref{fig:overview}
shows how the three fit together at inference.

The three networks are fitted in two stages. Stage~1
(Section~\ref{sec:autoencoder}) trains the encoder and the decoder together on
appearance alone, which fixes the latent chart; Stage~2
(Section~\ref{sec:lnn}) freezes them, encodes the clip once, and fits the
Lagrangian to the resulting trajectory, the order \citet{zhu2025cpae} follow.

\subsection{Stage 1: Autoencoder and Latent Chart}
\label{sec:autoencoder}

\paragraph{Continuity-preserving encoder.}
\label{sec:encoder}
What the encoder must deliver is regularity in time rather than accuracy, since
the dynamics objective (Section~\ref{sec:training}) reads $\dq$ and $\ddq$ from
the encoded trajectory, where a latent read off a pixel grid advances in the
staircase described in Section~\ref{sec:related-identify}. Neither depth nor
expressiveness is the remedy, the regularizer is~\cite{pottier2026plae}, and we
take the one of \citet{zhu2025cpae}, whose Lipschitz-constrained large filters
keep the latent state varying continuously with the underlying dynamics.
The latent width $d$ is read from the physics
when known (rigid-body DOFs, or the number of excited dynamic modes for a
deformable body), and otherwise set to the smallest value for which the
autoencoder reconstructs frames near-perfectly, in practice an upper bound on
the true count that a maximum-likelihood intrinsic-dimension estimate on the
encoded cloud~\cite{levina2004intrinsic} could replace.

\paragraph{Gaussian splatting decoder.}
\label{sec:decoder}
\label{sec:intuition}
\label{sec:schur}
Stacking the $T$ frames of a clip along a third axis yields a
\emph{space--time} volume $(x,y,t)$, which 4DGS~\cite{yang2024realtimephoto4dgs}
tiles with Gaussians: rendering frame $t$ is slicing that volume at time $t$.
Indexing on the clock is a choice, not a necessity: we tile the
\emph{space--state} volume $(x,y,\q)$ instead. A state the scene revisits is
then stored once rather than once per visit, and rendering follows whatever
$\q(t)$ the Euler-Lagrange ODE produces, from an unseen initial condition or
under an applied force.
Each of the $K$ Gaussians is placed directly in that volume, where it carries a
centre and a covariance that are fixed,
\begin{equation}
  \mu_k = \begin{pmatrix}\mu^{(s)}_k\\\mu^{(q)}_k\end{pmatrix}\in\R^{n+d},
  \quad
  \Sigma_k = L_k L_k^\top \in \R^{(n+d)\times(n+d)},
\end{equation}
with $L_k$ a lower-triangular Cholesky factor, plus scalar opacity $\alpha_k>0$
and colour $\mathbf{c}_k\in[0,1]^3$; the spatial dimension is $n=2$ for direct
video output (monocular training only) or $n=3$ for an interactive 3D scene with
novel views. All state dependence comes from conditioning on $\q$.
Partitioning $\Sigma_k$
into spatial ($s$, size $n$) and latent ($q$, size $d$) blocks, the rendered
$n$-D Gaussian is the Schur-complement conditional
\begin{align}
  \mu^{(s|\q)}_k &= \mu^{(s)}_k
    + \Sigma_{sq}\Sigma_{qq}^{-1}(\q - \mu^{(q)}_k),
    \label{eq:schur-mean}\\
  \Sigma^{(s|\q)}_k &= \Sigma_{ss} - \Sigma_{sq}\Sigma_{qq}^{-1}\Sigma_{qs},
    \label{eq:schur-cov}
\end{align}
with latent-weighted opacity
\begin{equation}
  \tilde\alpha_k(\q) = \alpha_k\cdot
    \exp\!\bigl(-\tfrac{1}{2}
      (\q-\mu^{(q)}_k)^\top\Sigma_{qq}^{-1}(\q-\mu^{(q)}_k)\bigr),
  \label{eq:latent_opacity}
\end{equation}
which suppresses Gaussians whose latent mean is far from $\q$: the conditional
centre shifts linearly in $\q$ through the cross-block $\Sigma_{sq}$, which
Section~\ref{sec:force} exploits, and the conditional covariance is
state-independent. The resulting $n$-D Gaussians are
rasterized with the \texttt{gsplat} CUDA kernel~\cite{ye2024gsplat}.
Appendix~\ref{sec:impl} gives the implementation details.
Appendix~\ref{sec:app-deform} compares this decoder, at equal budget, with the
alternative of $n$-D primitives warped by a deformation network, inspired by
\citet{wu20244dgs}.

\paragraph{Supervision.}
\label{sec:multiview}
Encoder and decoder are fitted jointly, on a photometric term and two
regularizers,
\begin{equation}
\begin{split}
  \mathcal{L}_\text{recon}
    = {}& \lambda_\text{L1}\mathcal{L}_\text{L1}
    + \lambda_\text{SSIM}\mathcal{L}_\text{SSIM}\\
    &+ \lambda_\text{nl}\mathcal{R}_\text{nl}
    + \lambda_\text{a}\mathcal{R}_\text{a},
\end{split}
  \label{eq:recon_loss}
\end{equation}
where $\mathcal{R}_\text{nl}$ is the nonlocal filter penalty that keeps the
encoder continuous~\cite{zhu2025cpae} and $\mathcal{R}_\text{a}$ caps the
aspect ratio of the rendered ellipsoids. Fitting the $(3{+}d)$D decoder from a
single view is ill-posed and calls for synchronized multi-view footage. Our
experiments substitute synthetic views for it: SHARP~\cite{mescheder2025sharp}
lifts every frame to a 3D reconstruction, novel views are rendered from each,
and the two photometric terms are evaluated on those as well.

\paragraph{Reading the chart.}
\label{sec:chart}
\label{sec:visibility}
Stage~1 closes by fixing, once and on the frozen networks, two quantities that
set how the latent chart is read from here on. The state itself is whitened
against its encoded statistics over the clip, so that Stage~2 sees a zero-mean,
unit-covariance coordinate (Appendix~\ref{sec:impl}); this is a change of chart,
not a regularizer. Discrepancies between states are weighed
differently. Latent directions are not
equally observable: some move many pixels, others almost none, and an isotropic
norm on $\q$ would weigh them alike, letting a direction nothing can see
dominate a fit or a score. The decoder supplies the right weighting, its
\emph{visibility} metric
\begin{equation}
  \bar G = \mathbb{E}_{\q}\!\left[
    \Bigl(\frac{\partial \Dec}{\partial\q}\Bigr)^{\!\top}
    \frac{\partial \Dec}{\partial\q}\right],
  \label{eq:visibility}
\end{equation}
the pull-back of the image metric averaged over the encoded trajectory. It is
strongly anisotropic in practice (Section~\ref{sec:exp-setup}); both the Stage~2
objective and our reported errors are taken in $\|\cdot\|_{\bar G}$.

\subsection{Stage 2: Latent Lagrangian Dynamics}
\label{sec:lnn}

With the autoencoder frozen, all frames are encoded once and a dynamics is
fitted to the resulting trajectory. The latent state evolves under an
Euler-Lagrange equation with dissipation,
\begin{equation}
  \frac{d}{dt}\frac{\partial\mathcal{L}}{\partial\dq}
  + \frac{\partial\mathcal{D}}{\partial\dq}
  - \frac{\partial\mathcal{L}}{\partial\q} = \f_\text{ext}(t),
  \label{eq:EL}
\end{equation}
where $\f_\text{ext}$ is zero during free evolution and carries the applied
forces otherwise, interactive or measured. $\mathcal{L}$ and $\mathcal{D}$ are neural
networks, differentiated automatically, with $\mathcal{L}=T-V$ split into a
kinetic and a potential term. The three paragraphs below constrain
$T$, $\mathcal{D}$ and $V$ in turn.

\paragraph{Inertia.}
\label{sec:force-flat}
The kinetic term is $T=\tfrac12\dq^\top M(\q)\dq$, with a learnable scalar mass
for $d=1$ and a learned SPD matrix $M(\q)=L(\q)L(\q)^\top+\varepsilon I$ for
$d>1$, whose Coriolis terms are derived from $\partial M/\partial\q$ rather than
modelled separately. $M$ depends on $\q$ because the photometric loss picks the chart and that chart
carries a scaling and a curvature of its own that the inertia must absorb. On
some systems the dependence is physical as well and survives any chart
(Section~\ref{sec:exp-krauss}). Learning
$M(\q)$ assumes nothing about how inertia is distributed over the scene. The
parameter-free alternative pulls the ambient metric back through the decoder
Jacobians $J_i=\Sigma_{sq,i}\Sigma_{qq,i}^{-1}$ of Eq.~\eqref{eq:schur-mean},
each primitive weighted by its presence at $\q$,
\begin{equation}
  M(\q) = m\sum_i \hat w_i(\q)\,J_i^\top J_i,
  \label{eq:latent-metric}
\end{equation}
whose state dependence comes from the presence weights $\hat w_i(\q)$ of
Eq.~\eqref{eq:latent_opacity}, so that it too carries Coriolis terms
(Appendix~\ref{sec:app-metric}). One mass $m$ is shared by every primitive: a
monocular video does not tell the Gaussians apart by material, so we take the
density to be uniform and let a primitive contribute inertia in proportion to
its presence and to nothing else. Its value is not fitted either, since it only
sets the global energy scale that Section~\ref{sec:force} leaves unidentified;
we fix it offline so that the initial natural frequency matches the observed
one. The form therefore carries no learned parameter and reuses the same
Jacobians that transport forces, at the price of two assumptions: that visible
material carries the inertia, and that it carries it uniformly.
Section~\ref{sec:exp-mass} settles the
choice experimentally, uniform density included; the results we report use the
learned mass.

\paragraph{Dissipation.}
The dissipation is $\mathcal{D}=\tfrac12\dq^\top C(\q)\dq$, with $C(\q)$ SPD
from a free $\q$-dependent Cholesky factor, built like the mass, and bounded
below by an isotropic floor $c_0$ whose role appears with the energy balance.

\paragraph{Potential.}
The potential $V$ carries the prior on the elastic landscape. An Input-Convex Neural
Network~\cite{amos2017icnn} is enough whenever a single convex basin is the
right assumption, as for our pendulum, rocking chair and hanging bag. It is too
strong for the soft continuum robot, whose elastic energy has crescent-shaped,
hence non-convex level sets around a still unique equilibrium, as already
observed on a cantilever beam~\citep{pottier2024lebnn}; we then compose the ICNN
with a learned diffeomorphism (an i-ResNet), which releases the convexity of the
level sets while preserving the unique stationary point, giving an invex
potential~\cite{sapkota2021invex}. Genuinely multi-well landscapes would call for a locally-convex
variant~\cite{jones2025multiwell} instead.

\paragraph{Measured actuation.}
\label{sec:pressure}
A logged actuation enters $\f_\text{ext}$ as a generalized force. For the
pneumatic robots we report on, the input is the vector of chamber pressures
$P\in\R^{n_c}$; with $\nu(\q)\in\R^{n_c}$ the latent chamber-volume map, the
virtual work $\delta W = P^\top(\partial\nu/\partial\q)\,\delta\q$ gives
\begin{equation}
  \f_P = b(\q)^\top P, \qquad b(\q) = \frac{\partial\nu}{\partial\q}\in\R^{n_c\times d}.
  \label{eq:pressure}
\end{equation}
We take $\nu$ concave, so that the loaded potential
$V_\text{eff}=V-P^\top\nu$ stays coercive under $P\ge0$ with a unique, stable
equilibrium.

\paragraph{Bounded and convergent response.}
The floor $c_0$ makes the energy strictly decreasing away from rest, so the free
dynamics stays bounded and settles at the single minimum of $V$, and at the
loaded equilibrium when a force is maintained, whatever the networks learn and
before any training (Appendix~\ref{sec:app-bound}). This is what turns an
under-constrained extrapolation into a plausible response to unseen forces:
lacking it, GaussianPrediction~\citep{zhao2024gaussianprediction} settles on a
sustained orbit, which no dissipative system admits
(Section~\ref{sec:exp-extrapolation}).

\paragraph{Fitting the dynamics.}
\label{sec:training}
We fit the dynamics one step at a time, as \citet{krauss2026von} do. From an
encoded state $(\q_i,\dq_i)$, one differentiable velocity-Verlet step of
Eq.~\eqref{eq:EL} produces $(\hat\q_{i+1},\hat{\dq}_{i+1})$, which is compared to
the next encoded state,
\begin{equation}
  \mathcal{L}_\text{dyn} =
  \bigl\|\hat\q_{i+1}-\q_{i+1}\bigr\|_A^2
  + \bigl\|\Delta t\,(\hat{\dq}_{i+1}-\dq_{i+1})\bigr\|_A^2 ,
  \label{eq:physics_loss}
\end{equation}
the velocity being counted in units of position per step. The norm is the one
that matters: to first order
$\Delta\q^\top\bar G\,\Delta\q \approx \|\Dec(\q+\Delta\q)-\Dec(\q)\|^2$, so
measuring in $\bar G$ is the decoded image error \citet{krauss2026von} minimize,
obtained without rasterizing anything, and it puts the weight on the coordinates
that actually move pixels. We do not use $\bar G$ bare: a barely visible
direction would be left unconstrained while possibly being stiff, and would
contaminate the visible ones through the off-diagonal coupling of $M$ and $C$,
so $A\propto\bar G+\rho\,\lambda_{\max}I$ caps the weighting ratio at
$1/\rho$, normalized to $\operatorname{tr}A=d$ so that $A=I$ recovers the plain
Euclidean loss. Minimizing the residual of Eq.~\eqref{eq:EL} directly, with
$\dq$ and $\ddq$ from a finite-difference stencil, is the classical alternative
\citep{cranmer2020lnn,friedl2025riemannian}; it is the same quantity before
integration, but it weighs every latent direction alike.

Both are single-step teacher forcing, hence the blind spot to long-horizon
stability recalled in Section~\ref{sec:related-mechanics}. We therefore select
models on multi-step rollout error over the training clip, rather than on the
single-step objective they minimize. Unrolling the prediction
over several symplectic steps~\cite{melchers2023comparison,chen2020symplectic}
is the alternative remedy, at an optimization cost we did not need to pay.

\subsection{Interaction and Inference}
\label{sec:force}

The map $\q\mapsto\mu_i(\q)$ that Stage~1 fits is a kinematics, and the same
Jacobian $J_i=\partial\mu_i/\partial\q$ that carries the inertia of
Eq.~\eqref{eq:latent-metric} carries the forces: a force applied to a primitive
does work on the latent state through $J_i$
(Appendix~\ref{sec:app-force}). That kinematics is already determined by the
autonomous video, so a loading never seen during training is nonetheless
prescribed by what the clip has shown. NeuROK~\cite{geng2025neurok} drives its
latent equation of motion through the same pullback, over deforming meshes
rather than splats.

At inference the user applies a force $\f\in\R^n$ at a scene point $x$
identified by raycasting. Its conversion to a latent generalized force reads
directly off the Schur mean (Eq.~\eqref{eq:schur-mean}): a primitive $i$
transports it as
\begin{equation}
  \f_\text{lat} = \left(\frac{\partial\mu^{(s|\q)}_i}{\partial\q}\right)^\top\!\!\f
  = J_i^\top\f
  = \bigl(\Sigma_{sq,i}\Sigma_{qq,i}^{-1}\bigr)^\top\f
  \;\in\R^d,
  \label{eq:force}
\end{equation}
added to the right-hand side of Eq.~\eqref{eq:EL} for as long as the force is
maintained; it is the canonical generalized force of virtual work, instantiated
with the learned kinematics (Appendix~\ref{sec:app-force}). The contact is
spread over a neighbourhood rather than attributed to the single closest
primitive: what enters Eq.~\eqref{eq:EL} is the convex combination
$\sum_i w_i(x)\,J_i^\top\f$, with weights
$w_i\propto\alpha_i\exp(-\|\mu^{(s|\q)}_i-x\|^2/2\sigma^2)$ normalized to one
over a brush of radius $\sigma$. Because $\phi_i:\q\mapsto\mu^{(s)}_i
+J_i(\q-\mu^{(q)}_i)$ is affine per primitive, each $J_i$ is constant and
already stored as a covariance block: the pull-back costs $O(nd)$, with no
network differentiation at inference.
\label{sec:force-scale}
What the kinematics cannot supply is the scale. The $J_i$ fix the direction and
the shape of the response, but the dynamics they feed is identified only up to
the rescaling
$(\mathcal{L},\mathcal{D})\to(\alpha\mathcal{L},\alpha\mathcal{D})$, which leaves
every observed trajectory unchanged. A force in physical units is therefore
defined up to a single energy factor $\kappa$, global to the model and fixed by
one measurement of a known force and its response, as announced in
Section~\ref{sec:intro}.

\paragraph{Interactive loop.}
\label{sec:inference}
Given one or two initial frames, we set $\q_0 = \Enc(I_0)$ and
$\dq_0 \approx (\q_1-\q_0)/\Delta t$ (or $\dq_0=0$ from a single frame) and
integrate Eq.~\eqref{eq:EL} with a symplectic scheme, which keeps the energy
from drifting over long free rollouts~\cite{chen2018neuralode}. Each interactive
step collects the user force, pulls it back through Eq.~\eqref{eq:force},
integrates with $\f_\text{ext}=\f_\text{lat}$ and renders: a force injected at
one step causally modifies all
subsequent states, and scaling $\mathcal{D}$, $T$ or $V$ here edits damping,
inertia or the effective potential of the simulated object. The loop runs on two
clocks: the dynamics, pull-back included, is integrated in real time at the $30$
or $60$\,Hz of the test case, and rendering is decoupled from it, above
$80$\,fps on a $16$\,GB GPU.

\section{Experiments}
\label{sec:experiments}

Four questions organise this section. Does the learned Lagrangian reproduce the
recorded motion and the physical quantities behind it
(Section~\ref{sec:exp-cases})? Does the mechanical prior buy anything past the
end of the clip, measured against a model that extrapolates the same video
without a Lagrangian structure (Section~\ref{sec:exp-extrapolation})? How does it
stand against published latent-dynamics models on a real, pressure-actuated soft
robot, and does its inertia have to be learned (Section~\ref{sec:exp-krauss})?
And does the force transport of Section~\ref{sec:force} behave as the geometry
predicts (Section~\ref{sec:exp-force})?

\subsection{Protocol}
\label{sec:exp-setup}

Every result below follows the two stages of Section~\ref{sec:problem}, with
nothing retrained end to end: the dynamics is always fitted on a latent chart
that the photometric stage has already frozen.

\paragraph{Systems.}
Every case is one or several monocular videos of a real object, filmed from a
single fixed viewpoint, with no state, force or torque annotation. All of them
are autonomous, left to move on their own once released, except the soft
continuum robot, which is driven throughout by its chambers and carries the only
sensor input of the set, the measured pressures. What separates these cases in
difficulty is first of all the latent dimension $d$, given in parentheses below
and set to the number of degrees of freedom the recording excites
(Section~\ref{sec:encoder}); they are listed here, and ordered in every table
that follows, by increasing $d$.
\emph{Pendulum} ($d=1$): a real pendulum from the Delfys75 dataset of
\citet{castaneda2025learning} at $60$\,fps, $632$ training frames and $633$
held out. \emph{Rainbow rocker} ($d=1$): a coloured wooden rocking toy at
$30$\,fps, one clip of $133$ frames of which $100$ train and the last $33$ are
held out; it is also the case we reconstruct in 3D, under multi-view
supervision, every other one being rendered in 2D. \emph{Rocking chair} ($d=1$): a
real chair oscillating on the floor at $30$\,fps, $330$ training frames and as many
held out. \emph{Hanging bag} ($d=2$): a backpack hung by its handle and left to
swing, filmed five times at $30$\,fps from the same viewpoint, four clips training the model ($2280$ frames) and the fifth ($270$)
held out entirely, so that its validation is an unseen recording rather than the
end of a seen one. \emph{Double pendulum} ($d=2$): a real chaotic double
pendulum at $24$\,fps, released near its downward equilibrium so that the
recorded swings stay small against the full $360^\circ$ range, $211$ training
frames and $53$ held out.
\emph{Soft continuum robot} \citep{krauss2026von}: a real pneumatic robot at
$60$\,fps in a one-segment ($d=2$, two pressure inputs, $43\,800$ training
frames) and a two-segment ($d=4$, four inputs, $43\,872$) configuration, driven by
measured chamber pressures, on the authors' own $80/20$ split; it carries
the only published baselines on identical data and is treated on its own in
Section~\ref{sec:exp-krauss}.

For the two systems that are not rigid bodies, the hanging bag and the robot,
$d$ is not read off a kinematic count but fixed empirically; how it compares
with the dimensions the published baselines need on the same recordings is
Section~\ref{sec:exp-krauss}.

\paragraph{Metrics.}
Pixel error alone conflates rendering, phase drift and damping, so we report
four quantities. The free-oscillation frequency error
$\varepsilon_{\mathrm{freq}}$ compares the oscillation frequency of a free
rollout to the one measured on the video itself, with the same windowed spectral
estimator on both. It is what tells whether the physics has been identified
rather than the frames merely fitted, a model being free to reproduce a clip
while carrying the wrong natural frequency. The latent error $\varepsilon_{\q}$ is an
NRMSE, where $1.0$ means no better than predicting the mean, and says how far
the predicted state drifts from the encoded one. The image error
$\varepsilon_{I}$ says how much of that drift is visible, and we report it as a
multiple of the AE floor $\varepsilon_{\mathrm{AE}}$, the static reconstruction
error of the frozen autoencoder: the floor measures the visual complexity of the
scene and of the renderer serving it, where the ratio of the two isolates what
the rollout itself costs. Both are averaged over two periods of the
system's own natural frequency, so that every case is scored over the same
amount of predicted motion. Latent norms are taken in the visibility metric
$\bar G$ of Eq.~\eqref{eq:visibility}, whose condition number reaches $341$ on
these systems, so an isotropic aggregation would not merely be noisier but
wrong. Section~\ref{sec:exp-krauss} adds the $0.5$\,s image MSE of
\citet{krauss2026von}, for comparability with their benchmark.

\paragraph{Reading the rollout figures.}
Figures~\ref{fig:rainbow-rollout}, \ref{fig:sac-rollout}, \ref{fig:dp-rollout}
and \ref{fig:krauss-npz-rollout} share one protocol. Each shows a
single rollout of the learned dynamics, integrated from the encoded state
of one frame, position and velocity, and never re-anchored afterwards.
The bottom row shows the real image, then the
render at the encoded $\q$, which is the renderer's own floor, then the
render at the predicted $\q$, each panel adding one error source to the
previous one. The insets magnify the window where the two renders differ most
in excess of what the autoencoder gets wrong there, so that what they
show is dynamics and not reconstruction, with an arrow giving the direction of
the measured shift. Latent components are
ordered by decreasing visibility, each panel carrying the share of
$\operatorname{tr}\bar G$ that coordinate holds: a large error on a barely
visible component degrades the predicted image only slightly.

\subsection{Results across test cases}
\label{sec:exp-cases}

\begin{figure}[tb]
\centering
\includegraphics[width=0.92\columnwidth]{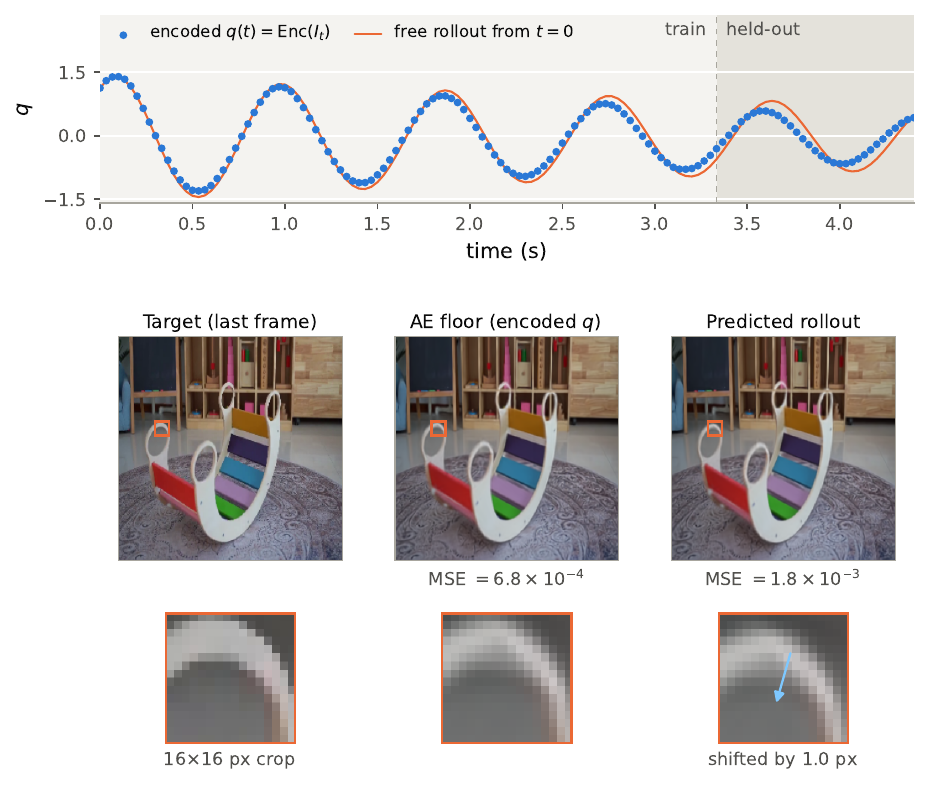}
\caption{%
  \textbf{One rollout across the train/held-out boundary} (rainbow rocker,
  $d=1$, the learned Lagrangian of Table~\ref{tab:physical}; protocol in
  Section~\ref{sec:exp-setup}). The shaded region is the held-out end of the
  clip, $33$ frames the model has never seen, over which the latent NRMSE is
  $0.335$. Five oscillations ahead, the handle edge is $1.0$ px from where the
  encoded state puts it, $2.4\times$ the local reconstruction error.}
\label{fig:rainbow-rollout}
\end{figure}

\begin{figure}[tb]
\centering
\includegraphics[width=0.82\columnwidth]{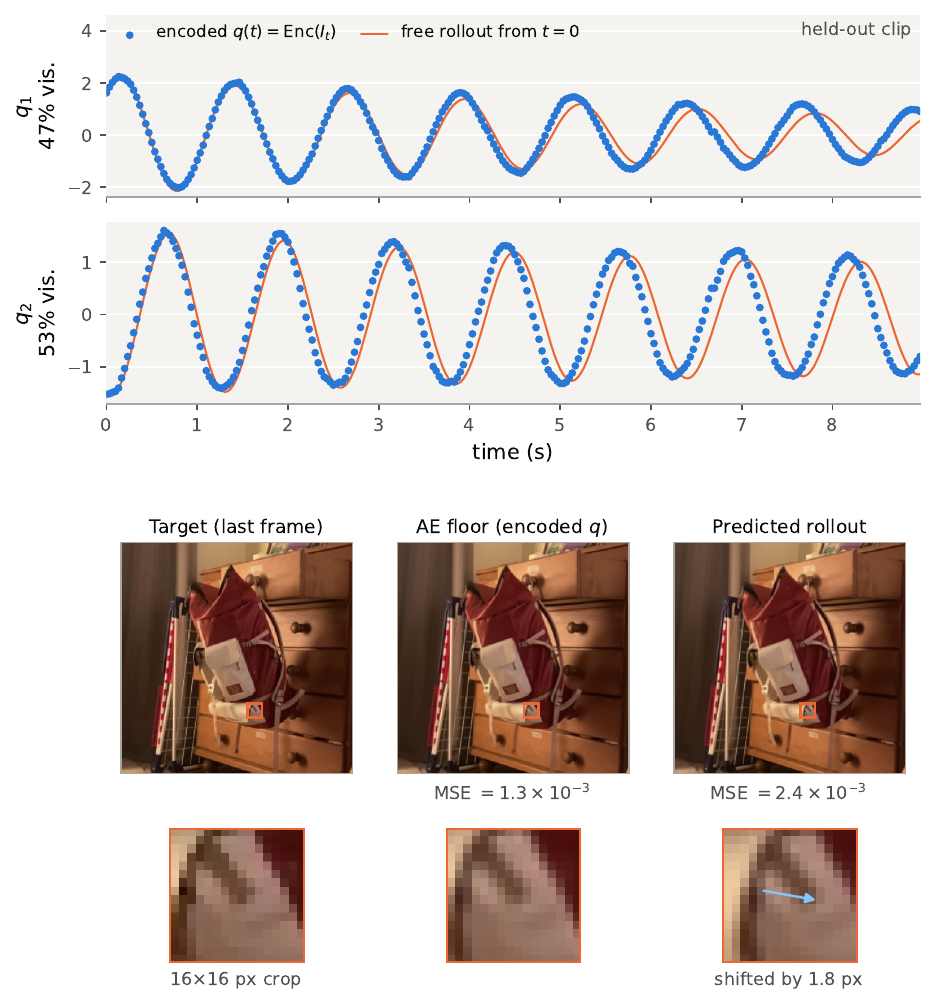}
\caption{%
  \textbf{Seven periods of free rollout on an unseen clip} (hanging bag,
  $d=2$), a separate recording, so the whole $9$\,s is held out. Amplitude is
  held and the residual error is a slow phase drift, $1.8$ px at the last frame,
  $4.0\times$ the local reconstruction error.}
\label{fig:sac-rollout}
\end{figure}

\begin{figure}[tb]
\centering
\includegraphics[width=0.82\columnwidth]{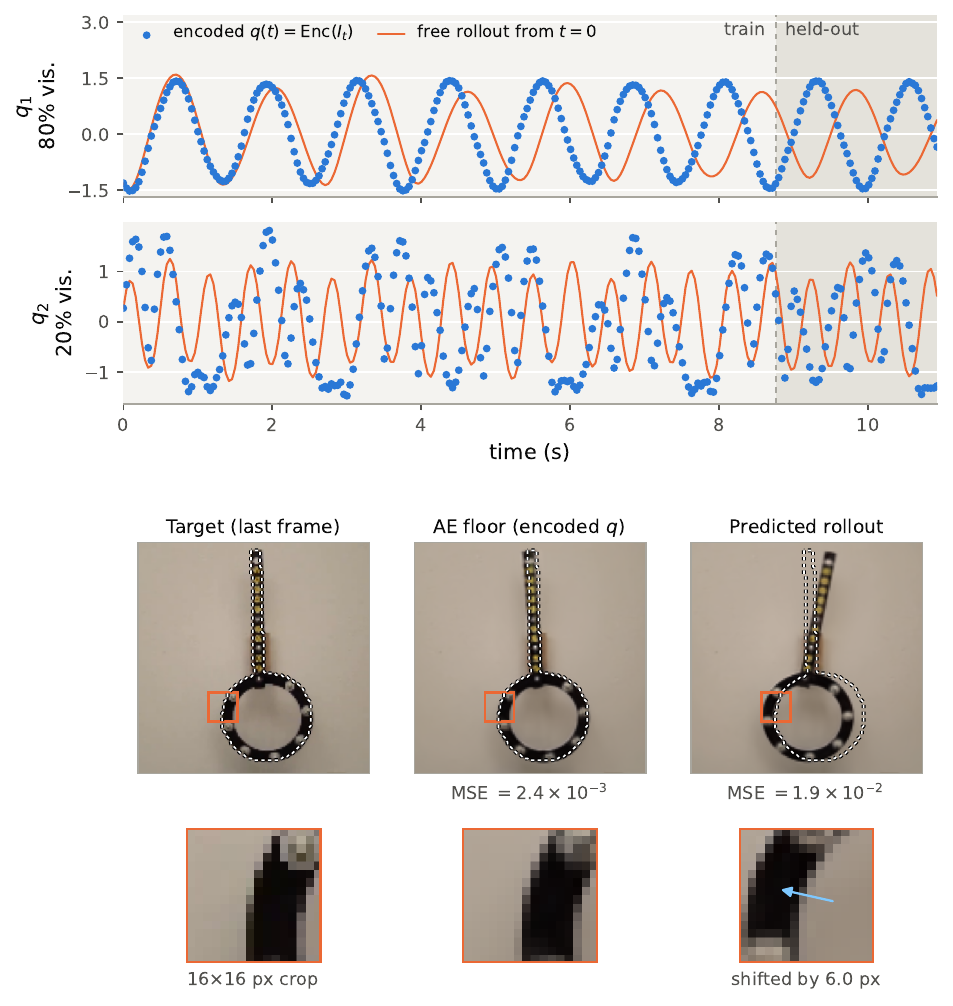}
\caption{%
  \textbf{Rollout on a chaotic system} (real double pendulum,
  $d=2$). The dominant coordinate is the slow swing of the assembly as a whole,
  and there the rollout holds amplitude and frequency, losing phase after two to
  three periods. The second carries the relative oscillation of the two arms,
  $\theta_2-\theta_1$, too chaotic to be identified from a clip this short: the
  model draws a regular oscillation where the encoded states scatter. The
  renderer is not what limits the horizon.}
\label{fig:dp-rollout}
\end{figure}

\begin{table}[h]
\centering
\caption{%
  \textbf{Free rollouts, scored on the four criteria of
  Section~\ref{sec:exp-setup}}, on the held-out split and with
  $\varepsilon_{\mathrm{AE}}$ in $10^{-4}$.}
\label{tab:physical}
\setlength{\tabcolsep}{3pt}
\small
\begin{tabular*}{\columnwidth}{@{\extracolsep{\fill}}lccccc@{}}
\toprule
System & $d$ & $\varepsilon_{\mathrm{freq}}$ (\%) & $\varepsilon_{\q}$ & $\varepsilon_{I}$ & $\varepsilon_{\mathrm{AE}}$ \\
\midrule
Pendulum              & 1 & $3.4$  & $0.308$ & $14.5$ & $0.93$ \\
Rainbow rocker        & 1 & $1.2$  & $0.125$ & $13.6$ & $0.35$ \\
Rocking chair         & 1 & $2.9$  & $0.230$ & $1.1$  & $1.56$ \\
Hanging bag           & 2 & $2.0$  & $0.150$ & $1.6$  & $0.93$ \\
Double pendulum       & 2 & $6.0$  & $0.947$ & $67.9$ & $1.39$ \\
Soft robot, 1 segment & 2 & $31.4$ & $0.299$ & $61.3$ & $0.12$ \\
Soft robot, 2 segments & 4 & $5.7$ & $0.273$ & $33.2$ & $0.63$ \\
\bottomrule
\end{tabular*}
\end{table}

On every system but the one-segment robot $\varepsilon_{\mathrm{freq}}$ stays
between $1$ and $6\%$: the natural frequency, which nothing supervises, comes
out of the learned dynamics to within a few percent of the one the video shows,
which is what \citet{castaneda2025learning} take as evidence that a model has
identified the physics. The one-segment robot is
the exception at $31.4\%$, and the cause is its recordings rather than our
model: they are made so as not to excite the natural mode, which nothing in them
therefore identifies (Section~\ref{sec:exp-krauss}). A frequency or a damping off by that little is
still enough to put the rollout out of step with the recording, which is what
$\varepsilon_{\q}$ registers over two periods
(Figs.~\ref{fig:rainbow-rollout} and \ref{fig:sac-rollout}), and the drift grows
with the horizon: on the hanging bag the latent error rises from $0.150$ over
two periods to $0.396$ over the seven of Fig.~\ref{fig:sac-rollout}. The double
pendulum is the case where it stays large, $\varepsilon_{\q}=0.947$, reaching
$1.544$ over the whole clip and moving the last frame by $6.0$ px, and what it
asks for is data: we identify it from the $211$ frames of one short clip, where the LNN
of \citet{cranmer2020lnn} learns the same chaotic system from $600\,000$
simulated states (Fig.~\ref{fig:dp-rollout}).

\subsection{What the Lagrangian prior buys}
\label{sec:exp-extrapolation}

\begin{figure}[h]
\centering
\includegraphics[width=\columnwidth]{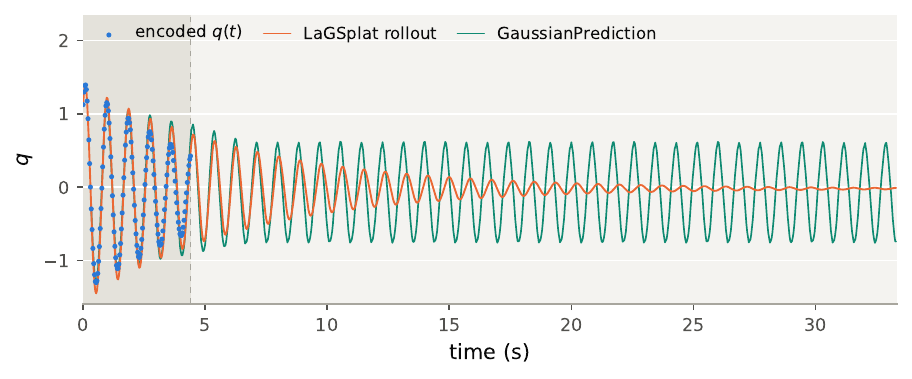}
\caption{%
  \textbf{Free extrapolation far past the clip} (rainbow rocker, $d=1$).
  The clip lasts $4.4$\,s, shaded; everything to its right is free
  extrapolation, with no ground truth to re-anchor either model. Both
  trajectories are read in the same latent chart, the
  baseline's predicted frames being passed through the frozen LaGSplat encoder.}
\label{fig:rainbow-equilibrium}
\end{figure}

To weigh what the mechanical prior buys we compare LaGSplat with
GaussianPrediction \citep{zhao2024gaussianprediction}, a member of the family of
Section~\ref{sec:related-identify} that carries none: Gaussian Splatting
forecast by a graph network over a few hundred key points, an evolution law it
is free to learn but that nothing requires to be dissipative whatever the
weights. Predicting motion rather than identifying a law covers a wider class of
extrapolation problems than ours. On a dissipative system that generality turns
into a weakness, because convergence to an equilibrium then has to be learned
where our model class cannot avoid it. Whether an unimposed dissipation is
learned nonetheless is what the comparison measures. We train it on the rainbow
rocker on the same split, then roll both models out far past the end of the
clip, into the regime where a dissipative system has long since come to rest.
GaussianPrediction exposes images and no state, so we read its predicted frames
through the frozen LaGSplat encoder to put both trajectories in one chart
(Fig.~\ref{fig:rainbow-equilibrium}). That reading only means something as long
as the encoder is fed images it can make sense of, and it is: long after the
motion they show has ceased to be that of a dissipative system, the predicted
frames are still visually coherent and still land within the latent range the
clip covers. What has degraded is the dynamics they display, not their
appearance, so the encoder is never asked to extrapolate.

\begin{table}[t]
\centering
\caption{%
  \textbf{Extrapolation past the training window} (rainbow rocker, $d=1$,
  train frames $0$--$99$, predicted frames $100$--$132$).
  PSNR is measured on the held out frames against the real video, LaGSplat autoencoder floor on them being $31.79$\,dB. The physics rows read both models and the real
  video in the latent chart of the frozen encoder: the frequency error is
  referenced to $1.1433$\,Hz and the dissipation coefficient is the envelope
  decay rate $1/\tau$ fitted over the filmed window ($\tau = 4.7$\,s), both
  measured on the encoded real frames. The last row comes from a free rollout of
  $1000$\,s, against a filmed clip of $4.43$\,s.}
\label{tab:extrapolation}
\setlength{\tabcolsep}{3pt}
\small
\begin{tabular*}{\columnwidth}{@{\extracolsep{\fill}}lcc@{}}
\toprule
                                 & LaGSplat & GaussianPred. \\
\midrule
\multicolumn{3}{@{}l}{\emph{Rendering} (PSNR, dB)} \\
Mean on held out frames    & $25.88$ & $\mathbf{27.86}$ \\
First predicted frame      & $27.28$ & $\mathbf{31.61}$ \\
Last predicted frame       & $23.53$ & $23.63$ \\
\midrule
\multicolumn{3}{@{}l}{\emph{Physics}} \\
Frequency error & $1.2\%$ & $0.9\%$ \\
Dissipation coefficient error & $\mathbf{29\%}$ & $60\%$ \\
Amplitude left after $1000$\,s & $\mathbf{0.00\%}$ & $53\%$ \\
\bottomrule
\end{tabular*}
\end{table}

On the held-out frames the baseline leads throughout
(Table~\ref{tab:extrapolation}): by $2.0$\,dB on average, by
$4.3$\,dB on the first predicted one, and by a margin too small to mean anything
on the last one. Little of
that is dynamics: it renders $200\,155$ Gaussians in its recommended setting
against our $15\,000$, and a one-frame misalignment is worth about $2$\,dB here,
so the whole rendering block sits within a frame of phase. GaussianPrediction
also gets the natural frequency of the system right, slightly better than we do
($0.9\%$ error against our $1.2\%$), and both models underestimate the damping,
on the training frames as on the held-out ones: fitted over the filmed window,
where all three series remain comparable, our dissipation coefficient falls
$29\%$ short of the one the real video shows and theirs $60\%$. Where the gap
really opens is the longer extrapolation. LaGSplat is structurally constrained to
converge to a stable equilibrium by its Lagrangian structure, whatever the
networks learn (Section~\ref{sec:lnn}, Appendix~\ref{sec:app-bound}), and that is
indeed what we observe, in a few tens of seconds. GaussianPrediction, on the
contrary, never converges: about $55\%$ of its initial amplitude is still there at
the end of Fig.~\ref{fig:rainbow-equilibrium}, $33$\,s in, and $53\%$ of it
after $1000$\,s, in what looks like a stable orbit and is entirely unrealistic for a
dissipative system like the rocker. Its renders stay photorealistic and inside
the observed latent range throughout, so its failure is dynamic and nothing
measured on the pixels reports it. The same failure
mode has been reported on gated recurrent cells fitted to dissipative structural
dynamics, where it motivated a cell whose internal energy cannot grow at
constant load and which is therefore structurally incapable of a sustained orbit
\citep{pottier2022gacrnn}. 

\subsection{Real actuated soft robot}
\label{sec:exp-krauss}

\paragraph{The system.}
The soft continuum robot of \citet{krauss2026von} is a real pneumatic actuator,
silicone segments bent by fibre-reinforced chambers and held to a plane, so that
a single camera sees the whole motion. Two configurations are released, one
segment driven by two independent pressures and two segments joined by a rigid
connector and driven by four, and both are driven continuously throughout the
recordings rather than released from rest.

How many coordinates that motion has is worth settling before the comparison. In
statics, the two pressures of a segment can lengthen it or bend it and nothing
else, so a segment holds two independent deformations, two for the first
configuration and four for the second. In dynamics, such a soft structure has in
theory infinitely many degrees of freedom, which are expressed or not depending
on the frequency content of the loading applied. The chambers are
driven between $0.04$ and $2$\,Hz \citep{krauss2026von}, while the robots
oscillate freely at $4.63$\,Hz on one segment and $1.566$\,Hz on two: the input
is too slow to reach the fast modes, and on one segment it barely reaches the
first one, at more than twice the top of that band, where the two-segment robot
sits inside it. Nothing beyond those two and four deformations was identifiable
in these recordings in our experiments, and the $32{\times}32$ frames make it
harder still, so we run at $d=2$ and $d=4$. The
authors use $d=6$ and $10$ instead, which follows from their dynamics rather
than from the data: their latent equation is linear, where two or four
coordinates can carry the whole motion only if the mass depends on $\q$, for a
reason that is in the object rather than in the latent chart. Bending or
compressing one segment brings its mass closer to the centre of rotation, which
changes its inertia in the corresponding direction of the latent space. No
constant matrix carries that, unless the dynamics is learned in a latent space
of higher dimension than the motion needs, which is what the extra coordinates
provide here.

\begin{figure}[h]
\centering
\includegraphics[width=\columnwidth]{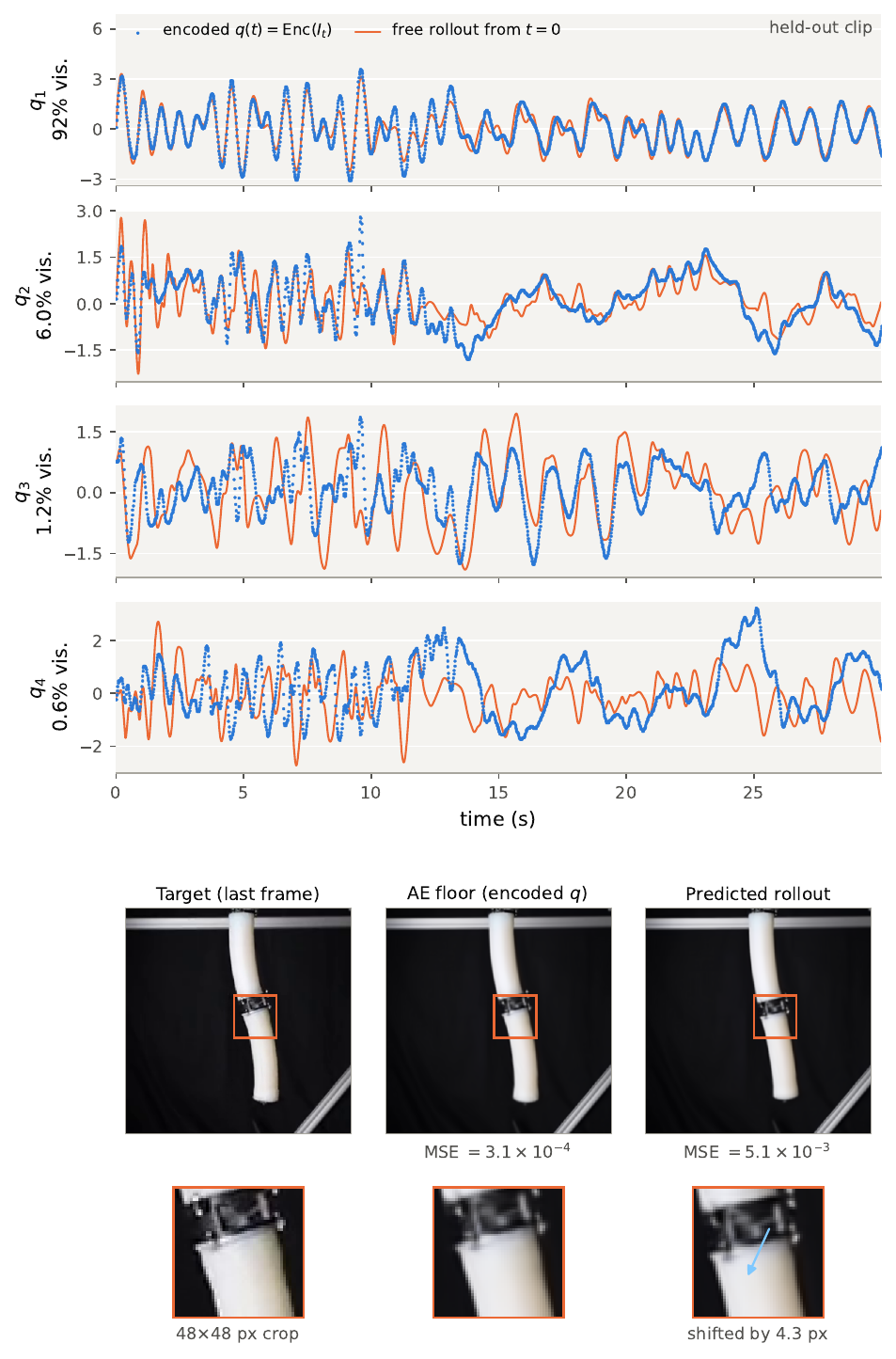}
\caption{%
  \textbf{Thirty seconds of pressure-driven rollout on the held-out split}
  (two-segment robot, $d=4$).
  Top: time evolution of the latent state $\q$ during the rollout, compared to
  the trajectory of the encoded ground-truth frames.
  Bottom: decoded frame at the last time step, rendered at $256\times256$.
  The first two components of $\q$ carry $98\%$ of the visible motion between
  them, which is why the offset between the prediction and the ground truth is
  only $4.3$ px although the last two components are poorly predicted.}
\label{fig:krauss-npz-rollout}
\end{figure}

LaGSplat handles higher resolutions easily, but we run on the authors' own
released data, their $32{\times}32$ frames at $60$\,fps, their pressure
alignment and their contiguous $80/20$ split, so that the comparison carries no
protocol difference. The code running this protocol end to end, from the
download of their data to our two-segment row of Table~\ref{tab:krauss}, is
public at \url{https://github.com/LouenPottier/LaGSplat}. Nothing in a Gaussian is
indexed by a pixel, so the same weights render at any resolution and only the
cost of rendering grows with the number of pixels, where the published decoders pay the change
of resolution in their own weights: the deconvolutional stack adds a stage and
doubles its channels each time the image doubles, and the attention decoder carries a
learned background map the size of the image, which would hold over $95\%$ of
its parameters at $256{\times}256$. That is the resolution
Fig.~\ref{fig:krauss-npz-rollout} is decoded at, retraining the decoder alone on
the frozen latent chart.
The measured chamber pressures enter as the generalised force $b(\q)^\top P$ of
Section~\ref{sec:pressure}: they are an input to the system rather than
supervision on it, the actuation matrix $b(\q)$ being learned like everything
else.

\paragraph{The baselines.}
Their study evaluates four latent-dynamics models under a common protocol, which
we adopt unchanged. They differ in their renderer and in
their dynamics, and we name each by that pair. The two dynamics are the
following, to be compared with our Eq.~\eqref{eq:EL},
\begin{align}
  \text{Osc.:}\ & M\ddot\q + D\dot\q + K(\q-\q_0) = g_\theta(P),
  \label{eq:base-osc}\\
  \text{Koopman:}\ & \begin{bmatrix}\q\\ \dot\q\end{bmatrix}_{t+1}
  = A\begin{bmatrix}\q\\ \dot\q\end{bmatrix}_{t} + h_\theta(P_t),
  \label{eq:base-koop}
\end{align}
where $M$ is diagonal, $K$ couples every coordinate to every other, $D=\alpha
M+\beta K$ is a Rayleigh dissipation, $A\in\R^{2d\times2d}$ is a free matrix, and $g_\theta$,
$h_\theta$ are small networks fed the last four pressure samples. Every
coefficient acting on the state is constant, so the pressure is the only place
where either model is nonlinear, and $A$ carries no time step: it is a map
between consecutive frames. Neither actuation term reads the configuration it
acts on, where our $b(\q)$ of Eq.~\eqref{eq:pressure} does. The two renderers
are as distinct. A deconvolutional decoder turns the latent vector into the
image through a stack of transposed convolutions, every pixel being produced
from the whole state and no part of the output attached to a coordinate. The
attention-broadcast decoder, which the authors abbreviate ABCD, expands each
latent coordinate over the pixel grid instead and gives it its own map, a
per-pixel weight obtained by a softmax across the maps and a learned background. Each map
therefore localises in the image what one coordinate moves, the background
holding the static part of the scene. The four share an encoder architecture,
each training its own autoencoder jointly with its own dynamics, which is why
every row carries its own AE floor. All four dynamics are linear in the latent state, so LaGSplat is the only
nonlinear Lagrangian in the comparison, and that is what their $d$ of $6$ and
$10$ pays for. The errors we report for them in
Table~\ref{tab:krauss} are their published ones, over the same $30$-step,
$0.5$\,s horizon and the same $50$ validation trajectories, rather than our own
re-measurements, except for the two one-segment models they leave unquantified;
their exact evaluation checkpoints are not recoverable from the release, and
re-running those checkpoints reproduces their published two-segment
figures to between $\times0.88$ and $\times1.04$ for every model but Osc.\,+\,ABCD, whose
released weights are the last epoch rather than the best ($\times2.11$).

Applying a force to the image, which is what the last column of
Table~\ref{tab:krauss} records, asks two things of a model in turn: the decoder
has to carry that force back to the latent coordinates, and the dynamics has to
turn the force it receives there into an acceleration. Only ABCD allows the
pull-back, and only at a handful of points: it attaches a position to what each
coordinate moves, the centre of mass of its mask, so a force can be applied at
those centres and nowhere else, $d/2$ of them once the coordinates are paired
into planar positions, where $J^\top$ places ours anywhere in the image. A
deconvolutional decoder produces every pixel from the whole state and exposes no
position at all. Only the oscillator has a place for that force
on the right-hand side of Eq.~\eqref{eq:base-osc}, where $M^{-1}$ turns
it into an acceleration and $K^{-1}$ into a shift of the equilibrium. A term added to
Eq.~\eqref{eq:base-koop} is only an increment of the state, no mechanical
reading being available from $A$, whose diagonal blocks leave the second half of
that state something other than the time derivative of the first. Osc.\,+\,deconv
therefore takes a force in latent coordinates but has nowhere in the image to
place it (${\sim}$), and the two Koopman rows cannot take one at all, whichever
decoder they carry ($\times$). Osc.\,+\,ABCD, which the authors call VON, is the
one baseline close to LaGSplat on this: it pairs the coordinates so that each
map carries a planar position and a square Jacobian, and the authors use it to
draw latent forces in the image plane $({\checkmark})$.

\begin{table}[t]
\centering
\caption{%
  \textbf{The soft continuum robot, against the four published baselines.}
  MSE in pixels of $[0,1]$; AE floor. $\varepsilon_{\mathrm{freq}}$, against the frequency of free oscillation measured on the depressurised tail of
  each clip, every baseline value being our own measurement on their released
  checkpoints.
  External force: $\times$, none; ${\sim}$, in latent coordinates
  only; $({\checkmark})$, in the image plane at $d/2$ points; $\checkmark$,
  anywhere in the image, through $J^\top$.}
\label{tab:krauss}
\setlength{\tabcolsep}{3pt}
\small
\begin{tabular*}{\columnwidth}{@{\extracolsep{\fill}}llccccc@{}}
\toprule
       &       &     & AE          & MSE          & $\varepsilon_{\mathrm{freq}}$ & Ext.  \\
Method & Dyn.  & $d$ & $(10^{-5})$ & $(10^{-3})$  & (\%) & force \\
\midrule
\multicolumn{7}{@{}l}{\emph{One segment}, two chambers} \\
Osc.\,+\,deconv     & lin.\    & $6$ & $1.03$          & $\mathbf{0.246}$   & $37.1$         & ${\sim}$ \\
Osc.\,+\,ABCD      & lin.\    & $6$ & $3.93$          & $0.574$            & $55.0$         & $({\checkmark})$ \\
Koopman\,+\,deconv  & lin.\    & $6$ & $1.15$          & $0.503$ & $10.3$         & $\times$ \\
Koopman\,+\,ABCD   & lin.\    & $6$ & $\mathbf{0.67}$ & $0.504$ & $\mathbf{8.8}$ & $\times$ \\
\textbf{LaGSplat} & nonlin.\ & $2$         & $1.25$          & $0.786$            & $31.4$         & $\checkmark$ \\
\midrule
\multicolumn{7}{@{}l}{\emph{Two segments}, four chambers} \\
Osc.\,+\,deconv     & lin.\    & $10$ & $5.43$          & $22.7$           & $76.8$         & ${\sim}$ \\
Osc.\,+\,ABCD      & lin.\    & $10$ & $14.5$          & $6.56$           & $31.4$         & $({\checkmark})$ \\
Koopman\,+\,deconv  & lin.\    & $10$ & $\mathbf{3.95}$ & $5.66$           & $18.3$         & $\times$ \\
Koopman\,+\,ABCD   & lin.\    & $10$ & $5.17$          & $0.984$          & $8.3$          & $\times$ \\
\textbf{LaGSplat} & nonlin.\ & $4$  & $6.31$          & $\mathbf{0.769}$ & $\mathbf{5.7}$ & $\checkmark$ \\
\bottomrule
\end{tabular*}
\end{table}

Excited well below its slowest mode, the one-segment robot
answers pressure with stiffness, its inertia contributing next to nothing, so
the recordings hold almost no evidence of the mass and the frequencies
identified there mean little, ours off by $31.4\%$. That figure averages the
whole relaxation: our rollout holds $4.61$\,Hz over its first half, then falls
$20$ to $24\%$ short as the motion dies out. Predicting $0.5$\,s amounts
to reproducing the quasi-static map from pressure to shape: the whole spread is
a factor $2.3$ there, against $23$ on two segments, and we come last on it. The
best of the four there is
Osc.\,+\,deconv, the simplest dynamics of the set, where both Koopman maps lead
on two segments. A dynamics with less freedom likely has less room to overfit
an inertia the recordings do not show.

Two segments is the harder of the two configurations, twice the coordinates
driven by twice the chambers and coupled through the connector, and every
baseline loses accuracy crossing to it where LaGSplat holds its own
($\times0.98$). It is there the most accurate model of the comparison, at
$0.769\times10^{-3}$, the median of five seeds and not the best of them, and it
gets there on $d{=}4$ coordinates where all four
baselines use $d{=}10$: a factor $8.5$ over Osc.\,+\,ABCD, the interpretable
oscillator that is our natural point of comparison, $30$ over Osc.\,+\,deconv,
$7.4$ over Koopman\,+\,deconv, and $1.28$ over Koopman\,+\,ABCD. Only that last
margin is thin, and it is held against the baseline that affords the least, the
one whose dynamics has no place for an applied force.

\paragraph{Decoder-induced inertia.}
\label{sec:exp-mass}
Section~\ref{sec:lnn} left open whether $M(\q)$ has to be learned at all.
Pulling the ambient metric back through the
decoder Jacobians (Eq.~\eqref{eq:latent-metric}) yields an inertia with no free
parameter, built from the very $J_i$ that already transport forces, which is how
NeuROK obtains its kinetic metric \citep{geng2025neurok}. It is state-dependent
although every $J_i$ is constant, since what varies along the trajectory is
which primitives are present: on the hanging bag it varies by a factor $2$ to
$3$ while staying well-conditioned, its condition number at most $3$.
Where the learned mass
leaves the distribution of inertia over the scene open, this one fixes it by
assuming a uniform density (Section~\ref{sec:lnn}), an assumption this
comparison is also a test of. We compare it with the
learned mass on the two-segment robot under the same frozen autoencoder, hence
the same latent chart and the same $J_i$, the same objective and the same
settings, so that the form of the mass is all that changes. The other terms keep
the structure of Section~\ref{sec:lnn} and are retrained, their weights free to
adapt to that form. The induced inertia is slightly behind
on both counts, $1.33\times10^{-3}$ over the $0.5$\,s horizon against the $0.769\times10^{-3}$ of
Table~\ref{tab:krauss}, and a frequency error of $6.9\%$ against its $5.7\%$.
The learned mass has more freedom, and that is what it buys; the residual gap is
what uniform density and the visible-material assumption cost, and it is small.
Carrying no free parameter at all, the induced inertia would still
rank ahead of three of the four baselines. A constant inertia, in contrast,
reaches $4.8\times10^{-3}$ on this test case: part of the state dependence
is in the object rather than in the chart, and no constant matrix carries it.
An inertia read off the decoder kinematics therefore accounts for the motion of
the robot several times better than the best constant one, although the map
$\q\mapsto\mu_i$ it is built from was fitted photometrically and never told
anything about mass. That map has learned a kinematics consistent with the way
the robot actually moves, which is what Section~\ref{sec:exp-force} relies on
when the same $J_i$ transport an applied force.

\subsection{Applying forces}
\label{sec:exp-force}

\begin{figure}[!t]
\centering
\includegraphics[width=\columnwidth]{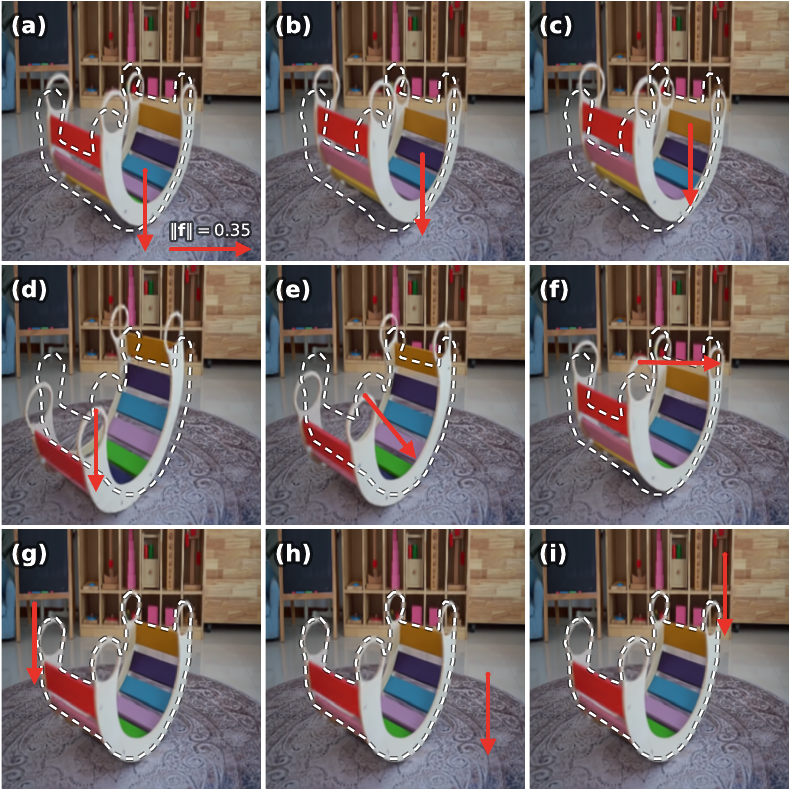}
\caption{%
  \textbf{Interactive force on the rainbow rocker ($d=1$).}
  The image-space force $\f$ is drawn as a red arrow at its application point,
  every panel applying the same magnitude on one arrow scale, so that only the
  application point and the direction change. The white dashed curve is the
  fixed rest envelope, the decode at $\q_\text{rest}=\arg\min V$.
  \textbf{(a)--(c), lever arm:} the same downward force applied at three
  slats progressively farther from the rocker contact. \textbf{(d)--(f),
  direction at a fixed material point:} the same point of the object pushed
  downwards, obliquely and horizontally.
  \textbf{(g)--(i), background:} the same force applied on the floor, the shelf
  and the rug.}
\label{fig:rainbow-force}
\end{figure}

\begin{figure}[!t]
\centering
\includegraphics[width=\columnwidth]{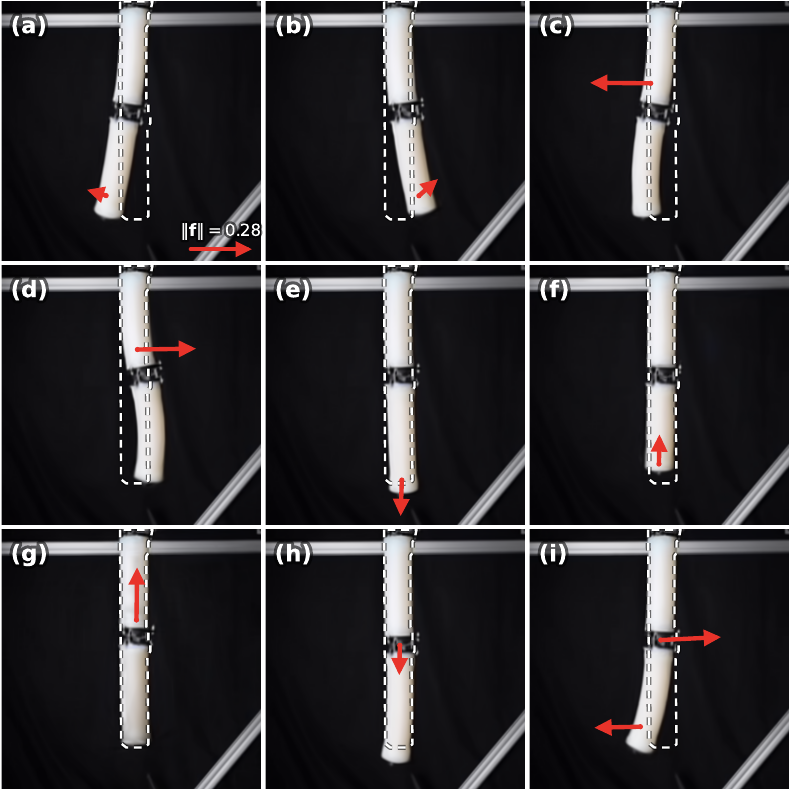}
\caption{%
  \textbf{Interactive force on the two-segment soft robot ($d=4$).}
  Nine equilibria of the pressure-driven continuum robot of
  \citet{krauss2026von}, all at one fixed operating pressure, so that what
  changes between panels is the applied force alone. The white dashed curve is
  the rest envelope at that operating point, and arrow length is proportional to
  $\|\f\|$ on one scale common to the nine panels. \textbf{(a),(b):} bending of both segments, left and right. \textbf{(c),(d):} bending of the first
 segment, left and right. \textbf{(e),(f):} force along the axis of the second segment, outwards then
  inwards, which stretches then shortens it. \textbf{(g),(h):} the same pair on
  the first segment. \textbf{(i):} two opposed forces, one at the junction
  between the segments and one at the free tip, under which the second segment
  bends while the first one nearly holds its rest envelope.}
\label{fig:krauss-force}
\end{figure}

The capability that separates LaGSplat from every entry of
Table~\ref{tab:positioning} is that a force applied on the image enters the
identified dynamics at inference, through $\f_\text{lat}=J^\top\!\f$, without
retraining and although no force was ever applied in the training video.
Figure~\ref{fig:rainbow-force} shows it live on the rainbow rocker and
Fig.~\ref{fig:krauss-force} on the two-segment soft robot. With the Lagrangian
and the dissipation of Section~\ref{sec:lnn}, the latent equation that receives
it is Eq.~\eqref{eq:EL} carrying both actuations,
\begin{equation}
  \frac{d}{dt}\frac{\partial\mathcal{L}}{\partial\dq}
  + \frac{\partial\mathcal{D}}{\partial\dq}
  - \frac{\partial\mathcal{L}}{\partial\q}
  = b(\q)^\top P + J(\q)^\top\!\f,
  \label{eq:forced-eom}
\end{equation}
the measured pressure of Eq.~\eqref{eq:pressure} and the applied force of
Eq.~\eqref{eq:force}, where $J(\q)^\top\!\f$ is the weighted sum of the
per-primitive transports $J_i^\top\!\f$ over the primitives under the click.
Training identifies the left-hand
side and the actuation map $b$; the force is added at inference, through
Jacobians the autonomous video has already fixed. That transport is a property
of the map $\q\mapsto\mu_i$ rather than of the coordinates used to write it:
writing the same kinematics in any other latent coordinates leaves every
rendered response of the two figures unchanged, and only the numerical values
of $\Delta\q$ depend on the chart the encoder selected
(Appendix~\ref{sec:app-force}). Both figures report equilibria
of that equation, each panel a live decode at the state $\q$ where the elastic
restoring force balances the right-hand side, the robot being held at one
operating pressure so that the applied force alone changes between panels. No
force ground truth exists for our systems, so we check predictions that follow
from the Jacobian structure and that the unidentified energy factor $\kappa$ of
Section~\ref{sec:force-scale} leaves invariant: ratios, signs and sums of
$\Delta\q$.

\paragraph{In the image plane.}
On the rainbow rocker ($d=1$), a force of fixed
magnitude applied at different points of the moving object must shift the latent
state by a $\Delta\q$ that grows with the lever arm of its application point and
reverses sign across the centre
of rotation; both halves hold, panels (a)--(c) of Fig.~\ref{fig:rainbow-force}
giving $\Delta\q=-0.50$, $-0.76$, $-0.84$ and panel~(d) reversing to $+1.22$ for
the same force on the other side. A force applied on a background Gaussian,
whose $J_i$ is null because it does not move with the object, must leave the
scene unchanged: panels (g)--(i) measure $|\Delta\q|\le0.007$, at least $70$
times below the on-object values, with no segmentation anywhere in the pipeline.
And the response to $\f_1+\f_2$ must equal the sum of the individual responses,
which no generative video model would satisfy: at a fixed application point this
is exact to $8\times10^{-8}$ relative, panels (d)--(f) predicting one force from
the other two by linearity. The point at which the arrow is drawn moves between
those three panels because the force follows the matter it was applied to: the
selection is frozen on the Gaussians at the click, and their image position is
re-evaluated at the deformed state.

\paragraph{On the soft robot.}
The kinematics is much richer here, the object being deformable and its latent
space four-dimensional, so the response to a force is no longer read off a
single lever arm. Panels (a) and (b) of Fig.~\ref{fig:krauss-force} give the
bending mode one would expect, both segments following the force to the left and
to the right. Panels (c) and (d) load the first segment alone, and the second
one bends although nothing is applied to it. Two readings are open and the
figure does not separate them: the learned kinematics may not decouple the
segments exactly, or the second segment, which hangs from the first and carries
its own weight, may bend under gravity once the first one has tilted. The axial
pairs (e)--(h) are consistent again: a force along the axis
stretches or shortens the segment it is applied to, without bending it and
without breaking the left-right symmetry, apart from a slight asymmetry at the
bottom of the second segment in (h). Panel (i) applies two opposed forces at the
two ends of the second segment, at the junction and at the free tip; the second
segment bends while the first one nearly holds its rest envelope, deflecting by
about a quarter of what the junction force alone produces in panel (c), and
against that force, toward the tip one. The sense is the expected one, the
distant force reaching the first segment as a moment while the weight of the
bent second segment shifts the same way; we read no static balance into the
amount, since gravity loads both segments and the figure cannot tell the two
contributions apart. Nothing in the model knows about
segments: the force is applied to whichever Gaussians the user selects, and the
response, whether it stays in one segment or reaches the other, comes from the
learned kinematics $\partial\mu/\partial\q$ alone. Turning these readings into a
quantified error would take a dataset in which known point forces are applied
and the resulting deformation is recorded, which none of the systems available
to us provides.

\paragraph{Forces in three dimensions.}
With the decoder supervised by several views, here the synthetic ones SHARP
renders from each lifted frame (Section~\ref{sec:multiview}) rather than a real
multi-camera capture, the same mechanism operates in the reconstructed
volume rather than in the image plane: a click casts a ray, the first Gaussian
it meets becomes the point of application, and $\f$ is a world-space vector at
that point. Panels (d)--(f) push the toy along a direction in which it never
moves in the training video, and nothing happens: the state stays where it was
to $|\Delta\q|<3\times10^{-9}$, an arrow that changes nothing. Such a force lies
in the kernel of the pullback. The toy has
a single degree of freedom, so each $J_i\in\R^{3\times1}$ has one column, and a
force orthogonal to it transports as $J^\top\!\f=-3.1\times10^{-17}$, machine
zero, whatever its magnitude. The points of the object do move vertically as
$\q$ varies, so the vertical is not in that kernel: a vertical force of the same
magnitude drives the toy from rest to a new loaded equilibrium at $\q=+1.52$,
in the expected direction.

\begin{figure}[!t]
\centering
\includegraphics[width=\columnwidth]{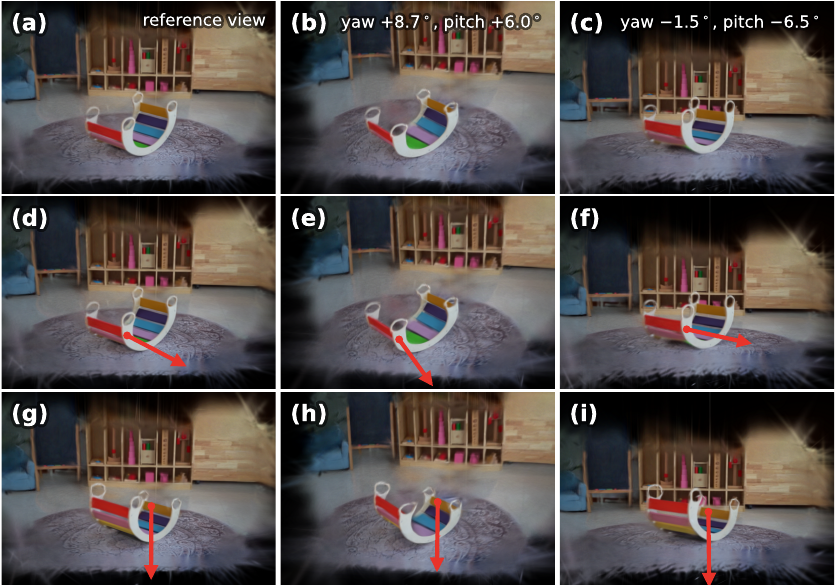}
\caption{%
  \textbf{Applying a force inside the 3D scene} (rainbow rocker, $d=1$,
  $15\,000$ Gaussians, decoder supervised by novel views). Each row is one
  state of the interactive simulation and each column one viewpoint of the
  free-viewpoint viewer, with the yaw and pitch offsets from the reference view
  written in the top row. $\f$ is drawn in red from its point of application,
  with the same magnitude everywhere and at a fixed length that only the
  viewpoint foreshortens, so arrow length carries no information.
  \textbf{(a)--(c):} the state at rest with no force at all, the static
  reference that replaces the dashed rest envelope of
  Fig.~\ref{fig:rainbow-force}. \textbf{(d)--(f):} a force orthogonal to
  $J_i=\partial\mu_i/\partial\q$ at the selected primitive, under which the
  state does not move: these renders differ from (a)--(c) only by the arrow
  overlay itself. \textbf{(g)--(i):} a vertical force, whose projection is
  nonzero, under which the toy settles at a new loaded equilibrium.}
\label{fig:rainbow-3d}
\end{figure}

\section{Discussion}
\label{sec:discussion}

\paragraph{Limitations.}
The Lagrangian prior narrows the class of systems the method applies to: a few
generalized coordinates, a smooth kinematics, and a dissipative dynamics that is
either autonomous or driven by a measured input. Contact, impact, changes of topology,
plasticity and hysteresis fall outside the dynamics we instantiate, where a
motion predictor carrying no mechanical structure covers them all. The boundary
is one of the dynamics we write, not of the representation it acts on: the
coordinates $\q$, the kinematics $\q\mapsto\mu(\q)$ and the pullback $J^\top$ do
not depend on the form given to the latent equation of motion. Any physics that
reduces to an equation of evolution on a small number of coordinates can
therefore take the place of the Lagrangian one and widen the class of systems
the method reaches, under one condition: that equation has to accept a
generalized effort on its right-hand side, which is what the pullback delivers.
A replacement that only predicts how $\q$ evolves, with nowhere for $J^\top f$
to enter, keeps the representation and loses the interaction. Contact and
plasticity both admit a form that meets the condition, and we leave them to the
directions below; neither is implemented here.

Learning from video adds a requirement of its own: the state has to be
observable in the images. A single viewpoint reads the motion the image plane exhibits, so
a displacement out of that plane would have to be supervised by real multiple
views. Plasticity raises the question more sharply, since a plastic variable is
not read off a frame the way a deformation is: two frames that coincide can
carry different loading histories. That does not put it out of reach, it puts it
out of sight of the per-frame encoder of Section~\ref{sec:encoder}. The
recording narrows the class further: only the
motions a video exhibits enter $\q$, so a degree of freedom the clip never
excites is one the model does not carry (Section~\ref{sec:intro}). The rainbow
rocker shows what that costs under force: a push orthogonal to its rocking plane
transports to machine zero and moves nothing
(Section~\ref{sec:exp-force}). That answer is right for a moderate push and
wrong for a large one, which would tip the toy over. The clip never shows it
tipping, so no coordinate carries that motion and the model stays rigid in that
direction whatever the magnitude. A state at which no Gaussian keeps any
presence renders nothing, so the Gaussians have to cover the whole region of $Q$
the state reaches at inference. The
frames, in turn, have to sample that region densely enough for the dynamics to
be identified over it. Both grow with the volume of the region, hence with $d$. One coordinate settles the question in a single
oscillation, which sweeps the whole segment it lives on; at $d=4$ a trajectory is
a curve in a four-dimensional space and covers a thin part of it, so the rest of
the coverage has to come from the input. The chambers of the soft robot are
driven throughout the recordings over a band of frequencies
(Section~\ref{sec:exp-krauss}), where the same robot released from rest would
have traced one decaying path and left the rest of $Q$ unvisited. For a system
with many more coordinates, the binding cost is therefore the recording itself,
long enough and driven variously enough to cover the space its state lives in.

The force mechanism, which is the central contribution, is never checked against
a measurement. Section~\ref{sec:exp-force} verifies predictions that follow from
the Jacobian structure, ratios, signs, sums and the kernel of the pullback, all
of them invariant under the energy factor $\kappa$; no error against a known
force is reported, because no dataset combines what that would take: one fixed
viewpoint, few degrees of freedom, and applied point forces recorded with the
resulting deformation. PokeFlex
\citep{obrist2024pokeflex} comes closest, real objects poked by a robot arm with
the interaction wrenches and the contact points logged, but its poking is
quasi-static, so it would test the loaded equilibrium and not the forced
dynamics. Data of that kind would nonetheless
settle the scale: $\kappa$ is global (Section~\ref{sec:force-scale}), so one
measured poke fixes it for the whole model, where video alone leaves it
undetermined. Every visible
primitive is treated as material of the same density. What that assumption costs
on the inertia was measured on one system only
(Section~\ref{sec:exp-mass}), where it is small, but a system whose parts differ
widely in density has no reason to be as forgiving. In the worst case a motion
in the image carries no mass at all: the primitives that follow a moving shadow still take a
share of it and still transport a force, so a push aimed at the shadow moves the
object.

Fitting appearance first and dynamics second is a bet. Against it,
\citet{friedl2025riemannian} report jointly trained latent Lagrangian models
outperforming a sequentially trained one by an order of magnitude on the decoded
prediction; for it, \citet{zhu2025cpae} make the latent continuous by
construction, fit the dynamics afterwards on a frozen encoder, and still beat
every jointly trained pairing of a standard autoencoder with a Neural ODE, an
HNN or a SympNet. The two are measured on different encoders, over simulated
states in one case and video in the other, and neither settles the question for
LaGSplat: fitting the same encoder and dynamics jointly, and comparing with the
two stages, is an experiment we have not run.

Finally, the latent dimension $d$ is chosen by hand, from the physics or from
the knee of the reconstruction error curve, a rule that can overestimate $d$, as
\citet{chen2022statevariables} report a reconstruction degrading while the
latent is still wider than the minimal set of state variables.

\paragraph{Future work.}
\textit{Measured forces.} Closing the loop on the force side asks for a
recording no dataset provides today: one fixed monocular viewpoint, a few degrees of freedom, and the
applied efforts logged with their point of application. Clips of human
manipulation are the natural case, and fall outside the class today because the
hand is an input the video never measures: masking the hand out of the frames and
logging the applied force and its contact point with an instrumented glove
\citep{liu2017glove,sundaram2019stag} would supply the missing record, and the
same measurement would fix the energy scale $\kappa$
(Section~\ref{sec:force-scale}) once for the whole model.
\textit{Contact and impact.} A differentiable contact model written on
$\q$~\cite{zhong2021contact} would resolve them in the latent space, grafted
onto a Lagrangian or Hamiltonian network and so keeping the right-hand side an
applied effort enters by, in place of the brush of Section~\ref{sec:force}, which injects an effort
but leaves the state continuous. Such a model needs a contact geometry, and the primitives are
already ellipsoids carried by the state: \citet{wang2026contactgaussian} take
the Gaussians as collision geometry and as appearance at once, and learn
rigid-body dynamics from video through a differentiable engine, which supplies
the geometric half of what a latent contact would take. That geometry constrains
the primitive itself: their Gaussians are spheres with frozen rotations, so that
a smoothed signed distance can be read off them, where
the latent axes we add live in the anisotropic covariance itself
(Section~\ref{sec:decoder}); recovering a contact geometry without giving up
that anisotropy is what remains open. The mechanical half is
the transport we already have, an impulse resolved on the primitives entering
the latent equation through the same $J^\top$.
\textit{Plasticity and hysteresis.} The other dynamics the Lagrangian form
leaves out, carried as internal variables beside $\q$ with a free energy and a
dissipation rate, in the manner of \citet{masi2022tann}, at the price of a larger $d$ and of
an encoder reading a window of frames rather than the single one of
Section~\ref{sec:encoder}. Those internal variables are compressed from
microscale fields a video does not provide. A video supervises the observed
response alone, which is what a recurrent constitutive model trains against,
carrying the loading history in a hidden state and never fitting the internal
variable itself \citep{gorji2020rnn}.
\textit{Coupling with other simulators.} Rigid bodies, a finite element
structure or a particle solver, placed beside the identified object in one
virtual environment: a reaction
computed by the environment enters the latent equation through $J^\top$, and the
primitives $\mu_i(\q)$ return the geometry the environment collides against, in
both directions at inference and without retraining. VR-GS
\citep{jiang2024vrgs} builds the interactive setting with the physics prescribed
on a tetrahedral cage around the Gaussians rather than identified from the
video; what would change here is that the object carries its own identified
dynamics into the scene.
\textit{Control from video.} The identification returns a mass, a dissipation, a
potential and an actuation matrix $b(\q)$, which is what model-based control
asks for, so a controller could be designed on the identified model of a soft
robot with no joint encoder ever attached to
it~\cite{lutter2019delan,kotecha2025lnnquadruped}.
\textit{Designed excitation.} Which input sequence, or which set of releases,
covers $Q$ best for a given number of frames is a question classical system
identification treats under the name of persistent excitation.
\textit{Priors across systems.} Learned once rather than once per object, so
that the coverage above is not identified from scratch every time. The two branches of Sections~\ref{sec:related-generative}
and~\ref{sec:related-identify} sit on either side of what this would take: one
learns over many systems what motion looks like and never registers it against
the particular object filmed, the other identifies one object from its own video
and starts from nothing every time. Between the two, a prior over the kinematics
and over the dynamics itself, the mass, the potential and the dissipation,
amortised over many systems and then fitted to one, would let a short clip
identify what a long one identifies today, in the spirit
of~\cite{geng2025neurok,xiao2026lawm}.
\textit{A chart shared across systems.} Such a prior has to be written on a
chart, and the transported effort does not
depend on the one the encoder selects (Appendix~\ref{sec:app-force}): a single
latent space can therefore carry every system, what distinguishes them being the
kinematics $J_i$ and the functions $M$, $V$, $C$ and $b$ written on it rather
than the coordinates themselves. The counterpart is that the coordinates of two
systems are then unrelated, so a prior learned on one says nothing about the
other, and a shared chart needs a rule giving each coordinate the same role
everywhere. NeuROK obtains one by training a single encoder over many objects,
so that they are all read into the same latent space \cite{geng2025neurok}. A
general-purpose visual encoder is shared in that sense already: DINOv2
\cite{oquab2023dinov2} and JEPA-style video encoders \cite{wang2026adajepa} map
every scene into one space, and using them frozen would also remove the
per-scene encoder training our pipeline needs. What is not settled is whether
their features resolve the few degrees of freedom that carry the motion, which
is what $\q$ has to be.
\textit{Several systems in one clip.} Each of our clips films a single object. Two
objects moving at once should be carried by coordinates that separate, disjoint
blocks of $\q$ with a mass and a potential that do not couple them, and the force
mechanism supplies the test: a push on one leaves the other still. An exactly
identified model would pass it whatever the coordinates, the equations being
covariant, so what is at stake is the fit rather than the principle, the block
structure being a prior that nothing in the losses currently rewards; and the
joint state space is the product of the two, which brings back the coverage
question above.
\textit{Per-primitive density.} The decoder-induced metric assumes it uniform
for want of a way to tell the primitives apart by material
(Section~\ref{sec:lnn}): a network predicting a density from appearance, as
\citet{le2025pixie} predicts material fields and \citet{lin2025omniphysgs}
constitutive models, would supply the $m_i$ that Eq.~\eqref{eq:latent-metric}
shares between all primitives today, fixing their ratios though not the global
scale $\kappa$. Telling the primitives apart by material also means telling
apart those that carry no material at all: segmenting the object from the shadow
it casts would drop the shadow primitives from both sums, the mass of
Eq.~\eqref{eq:latent-metric} and the brush of Section~\ref{sec:force}, where
they currently weigh and take a push like the rest.
\textit{Articulated systems and constraints.} Multi-body systems via a
Lagrangian graph network~\cite{bhattoo2022lagrangian}, and systems with explicit
constraints, which LaGSplat does not enforce~\cite{finzi2020simplifying},
trading that guarantee for the ability to handle an unknown constraint
topology.

\section{Conclusion}
\label{sec:conclusion}

We presented \textbf{LaGSplat}, which replaces the time parameterization of
dynamic Gaussian Splatting by a learned physical state $\q$: one
low-dimensional vector is at once the generalized coordinate of a dissipative
Lagrangian identified from video and the conditioning variable of a decoder
whose primitives are explicit points moving with the object. Nothing supervises
$\q$: the training signal is the video itself.

Across seven real systems, from one degree of freedom to four, the natural
frequency comes out of the learned dynamics within a few percent of the one the
video shows, on every recording that excites it. On the two-segment soft
continuum robot of \citet{krauss2026von}, evaluated on the authors' own data,
split and protocol, LaGSplat is the most accurate of the five models compared,
with four coordinates where the published baselines use ten; an inertia read off
the decoder kinematics, carrying no free parameter at all, still ranks ahead of
three of them. Against a motion predictor with no mechanical structure,
GaussianPrediction \citep{zhao2024gaussianprediction}, it
settles at a stable equilibrium the training clip never shows, where the
baseline is still orbiting a thousand seconds later.

The capability that motivates the construction is force at inference. A force
applied anywhere in the image, or in the reconstructed volume under multi-view
supervision, transports through the decoder Jacobians into the identified
equations of motion, without retraining and although no force was ever applied
in the training video. Its direction and its shape are fixed by the kinematics
that the autonomous video has already determined; its amplitude is fixed only up
to one global energy factor.

The Lagrangian prior narrows the class of systems the method reaches, and buys
inside it a response to unseen forces that stays bounded and physically
plausible.

\section*{Code availability}
The code and the protocol reproducing the two-segment row of
Table~\ref{tab:krauss} are available at
\url{https://github.com/LouenPottier/LaGSplat}.

\section*{Acknowledgements}
The author thanks Mathis Boston and Jérémie Le Garrec for the many discussions
on the Gaussian Splatting part of the pipeline.

\appendix
\section{Derivations}
\label{sec:app-derivations}

This appendix collects the derivations relegated from Section~\ref{sec:method}:
what the learned kinematics transports, forces and inertia alike
(Section~\ref{sec:app-force}), and the energy balance behind the
bounded-response remark (Section~\ref{sec:app-bound}).

\subsection{Forces and inertia from the learned kinematics}
\label{sec:app-force}

Stage~1 gives each primitive $i$ a forward kinematics, the map from the
configuration $\q\in\R^d$ to the centre of that primitive in the $n$-D render
space,
\begin{equation}
  \phi_i(\q) = \mu^{(s)}_i + J_i\,(\q-\mu^{(q)}_i),
  \qquad J_i=\Sigma_{sq,i}\Sigma_{qq,i}^{-1},
\end{equation}
affine by Eq.~\eqref{eq:schur-mean}. It plays the role of the position
$\mathbf r_i(\q)$ of particle $i$ in analytical
mechanics~\citep{goldstein2002classical}. Once that map is fixed, both the generalized force
and the kinetic energy follow from it by the textbook rules, with no further
modelling choice.

\paragraph{Transport of a force.}
A virtual displacement $\delta\q$ moves the primitive by
$\delta\mu_i=J_i\,\delta\q$, so a force $\f$ applied to it does the work
\begin{equation}
\begin{split}
  \delta W &= \f^\top\delta\mu_i = \f^\top J_i\,\delta\q\\
           &= (J_i^\top\f)^\top\delta\q,\qquad\forall\,\delta\q .
\end{split}
\end{equation}
The latent force we are after is the vector $\f_\text{lat}$ that does this same
work against any virtual displacement of the latent space, that is
$\delta W=\f_\text{lat}^\top\delta\q$. The last line has exactly that form, with
$\f_\text{lat}=J_i^\top\f$, and no other vector does, since two vectors with the
same product against every $\delta\q$ are equal.
Equation~\eqref{eq:force} is thus the canonical generalized force
$Q_j=\sum_i\mathbf F_i\!\cdot\!\partial\mathbf r_i/\partial q_j$ evaluated on
the learned kinematics, which is what makes its injection into
Eq.~\eqref{eq:EL} consistent with the energy balance of
Section~\ref{sec:app-bound}.

\paragraph{Invariance to the latent chart.}
The latent coordinates come out of the encoder and the whitening that follows
it, so they carry no physical meaning and are fixed only up to a smooth change
of variables $\q=\psi(\tilde\q)$. Writing $\Psi=\partial\psi/\partial\tilde\q$,
the chain rule replaces $J_i$ by $J_i\Psi$, hence the transported force by
$\Psi^\top\f_\text{lat}$; a virtual displacement of the new chart is
$\delta\q=\Psi\,\delta\tilde\q$ in the old one. The two factors cancel and the
work is the same on both sides,
\begin{equation}
  (\Psi^\top\f_\text{lat})^\top\delta\tilde\q
  = \f_\text{lat}^\top\Psi\,\delta\tilde\q
  = \f_\text{lat}^\top\delta\q .
\end{equation}
The response to a given user force therefore does not depend on the chart the
encoder happened to select.

\paragraph{The same map carries the inertia.}
\label{sec:app-metric}
Giving every primitive the same mass $m$ and weighting it by its presence
$\hat w_i(\q)$, the latent opacity of Eq.~\eqref{eq:latent_opacity} normalized
over primitives, the kinetic energy of the material points reads
\begin{equation}
  T=\tfrac12\,m\sum_i \hat w_i(\q)\,\|\dot\mu_i\|^2
   =\tfrac12\,\dq^\top\Bigl(m\sum_i \hat w_i(\q)\,J_i^\top J_i\Bigr)\dq
\end{equation}
because $\dot\mu_i=J_i\dq$, which is the mass matrix of
Eq.~\eqref{eq:latent-metric}. All the state dependence sits in the gating
$\hat w_i(\q)$, which is why $M(\q)$ varies along a trajectory, and carries
Coriolis terms, although every $J_i$ is constant.

\subsection{Bounded forced response}
\label{sec:app-bound}

We record the energy balance behind the bounded-response remark of
Section~\ref{sec:lnn}, under the assumptions realized by the architecture:
(A1)~$M(\q)=L_m L_m^\top+\varepsilon I\succ0$ SPD, so $T\ge\tfrac{\varepsilon}{2}\|\dq\|^2$;
(A2)~$V$ coercive (radially unbounded), enforced by a quadratic floor
$+\tfrac{\varepsilon_V}{2}\|\q\|^2$, without which the energy would still
decrease while its sublevel sets stayed unbounded, leaving $\q$ free to drift;
(A3)~$C(\q)=L_cL_c^\top+c_0I\succeq c_0I$, with the isotropic floor $c_0>0$ of
Section~\ref{sec:lnn}.

With $\mathcal{L}=T-V$, $\mathcal{D}=\tfrac12\dq^\top C(\q)\dq$ and total energy
$E=T+V$, differentiating along a solution of Eq.~\eqref{eq:EL} and using Euler's
theorem for the quadratic $\mathcal{D}$
($\dq^\top\partial\mathcal{D}/\partial\dq=2\mathcal{D}$) gives
\begin{equation}
  \dot E = \f_\text{ext}^\top\dq \;-\; \dq^\top C(\q)\,\dq
        \;\le\; \f_\text{ext}^\top\dq \;-\; c_0\|\dq\|^2 .
  \label{eq:energy-balance}
\end{equation}

\paragraph{Free rollout.}
With $\f_\text{ext}=0$ the balance reads $\dot E\le-c_0\|\dq\|^2$, so $E$ is a
Lyapunov function, strictly decreasing outside rest, and the trajectory stays in
the sublevel set $\{E\le E(0)\}$, which (A1) makes bounded in $\dq$ and (A2)
in $\q$. By LaSalle's invariance principle~\citep{khalil2002nonlinear} it
converges to a state where the velocity is zero and stays zero, which by
Eq.~\eqref{eq:EL} requires $\partial V/\partial\q=0$.
Both potentials of Section~\ref{sec:lnn} have exactly
one such point, their global minimum, the ICNN by strong convexity under the floor of (A2)
and the invex form by construction. Every free trajectory therefore comes to
rest at that single state, from any initial condition and however far a force
had driven it: sustained oscillation is excluded by the architecture, before any
training and whatever the networks have learned.

\paragraph{Maintained force.}
This is the regime of Section~\ref{sec:exp-force}, where a force is held while
the state settles. For a constant $\f_\text{ext}=\f$, the same differentiation
applied to the loaded energy $E_\f=T+V-\f^\top\q$ cancels the injected power
exactly,
\begin{equation}
  \dot E_\f = \dot E - \f^\top\dq = -\,\dq^\top C(\q)\,\dq \;\le\; -c_0\|\dq\|^2 ,
\end{equation}
and $E_\f$ is coercive as well, the quadratic floor of (A2) dominating the
linear term $\f^\top\q$. The free-rollout argument therefore applies verbatim
with $V$ replaced by $V-\f^\top\q$: the response stays bounded and converges to
a loaded equilibrium $\partial V/\partial\q=\f$, which is what the panels of
Figs.~\ref{fig:rainbow-force} and~\ref{fig:krauss-force} report.

Those statements are continuous-time; the symplectic scheme of
Section~\ref{sec:inference} is what keeps a long discrete rollout from drifting
away from them. They bound the response; they do not make it quantitatively
exact far from the training data.

\section{Deformation-Field Decoder at Equal Budget}
\label{sec:app-deform}

Section~\ref{sec:related-rendering} leaves two ways of giving a primitive a
position that follows the state. Ours places the primitive in the $(n{+}d)$D
volume and conditions it on $\q$ (Section~\ref{sec:decoder}), which makes
$\q\mapsto\mu_i$ affine with a constant Jacobian. The other keeps the primitive
$n$-D and warps it with a network of $\q$, after \citet{wu20244dgs}.

The second family contains the first. Its primitives follow whatever curve the
network draws, so each of them can in principle stay on one piece of the object
over the whole motion, which a straight line through the latent volume cannot
do. That freedom buys no image quality here, and the primitives stop following
the matter they render.

Eq.~\eqref{eq:recon_loss} has no term that holds a primitive on the same piece
of matter. The loss scores rendered frames, and a frame is explained as well by
a primitive that has moved with the object as by another that has taken its
place. Material tracking is therefore never supervised. It comes out of the
fit: the same primitives have to explain every frame of the clip, and
between two nearby states the cheapest way for one of them to do so is to shift
slightly and keep covering what it already covered, rather than leave it to a
primitive arriving from elsewhere. In our decoder a
visible primitive translates along a line in $\q$, at a rate fixed once and for
all by the block stored with it, and the latent gate of
Eq.~\eqref{eq:latent_opacity} fades it in and out along the way without taking
it off that line. The deformation field ties it to no trajectory: centre and
extent are free functions of $\q$, so one primitive shrinking below the pixel
while another grows in its place costs nothing.

We ran both on the rainbow rocker with $15\,000$ primitives on each side, the
same photometric loss and regularizers, the same optimizer and schedule, so that
the parameterization of $\q\mapsto\mu_i$ is the only difference. The deformation
side uses $9.8$M parameters, ours $195$k.

\begin{figure}[t]
\centering
\includegraphics[width=\columnwidth]{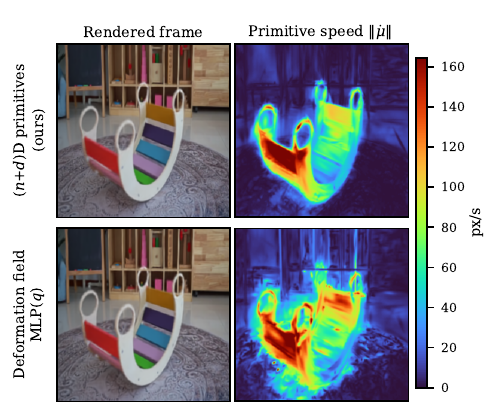}
\caption{%
  \textbf{The same motion read off the two decoders} (rainbow rocker, $d=1$). Each
  primitive is colored with the norm of its velocity
  $\|\dot\mu_i\|=\|J_i\dq\|$ and rasterized by the same kernel, with the same
  opacities and covariances as the frame beside it.}
\label{fig:deform-velocity}
\end{figure}

\paragraph{Reconstruction.}
At the end of training the deformation field is behind on every term of
Eq.~\eqref{eq:recon_loss}: L1 $0.0157$ against $0.0142$, the
SSIM term $0.076$ against $0.063$. Both
decoders are scored on the frames they were fitted on, so this compares capacity
spent and not generalization, and at this budget the extra $9.6$M weights buy no
image quality.

\paragraph{The Jacobian field.}
Primitive velocities are not supervised either. Figure~\ref{fig:deform-velocity}
reads $\dot\mu_i=J_i\dq$ off both decoders at the same state. The two renderings
are hard to tell apart, the two velocity fields are not. Ours varies smoothly
over the rocker; the deformation field is rough, a primitive differing from its
immediate neighbours by $6.9$ px/s in median against $1.4$ px/s for ours,
and it sets primitives in motion over the static background and over the rocker
itself. A decoder can therefore render the clip well and still leave
$\partial\mu_i/\partial\q$ meaningless, which is the derivative that
Eq.~\eqref{eq:force} pulls an applied force back through and that
Eq.~\eqref{eq:latent-metric} reads the inertia off.

\section{Implementation Details}
\label{sec:impl}

The same architectures are used on every system. The latent dimension $d$, the image
resolution, the number of primitives and the length of the runs are set per
system.

\paragraph{Architectures.}
The encoder is a continuity-preserving convolutional
autoencoder~\cite{zhu2025cpae}: three convolutions with $12{\times}12$ filters,
two channels each, stride $2$ and ELU activations, average-pooled to
$4{\times}4$ and mapped to $\q$ by one linear layer, about $2\,000$
parameters. The decoder holds $K$ Gaussians of dimension $n+d$, $1\,000$ to
$15\,000$ of them depending on the scene, which is $10^4$ to
$2\times10^5$ parameters. The LNN is under $15\,000$ parameters: a scalar
ICNN~\cite{amos2017icnn} for the potential, hidden $[64,64]$ with softplus
activations, composed with a $2$-block i-ResNet diffeomorphism (width $64$,
Lipschitz cap $0.9$) when an invex potential is required
(Section~\ref{sec:lnn}); an MLP with hidden $[64,64]$ for the Cholesky factor of
$M(\q)$, SPD floor $\varepsilon=0.1$; another for the full $C(\q)$, initialised
at $0.2\,I$ with isotropic floor $c_0=1$; and one invex ICNN head per chamber
for the actuation map $\nu(\q)$.

\paragraph{Training.}
Both stages are trained with Adam. Stage~1 trains encoder and decoder jointly
for about $5\times10^4$ optimizer steps, the number of epochs following from the
size of the clip and the period of the Gaussian lifecycle scaled with it, on
$0.08\,\text{L1}+0.02\,(1-\text{SSIM})+0.01\,\text{anisotropy}$, with a
$7{\times}7$ SSIM window; a Gaussian whose peak opacity falls below $0.02$ is
periodically recycled, either by splitting an overloaded alive Gaussian along
the principal eigenvector of its full $(n{+}d)$D covariance or by teleporting it
to a high-error region. Latent whitening $\q_\text{white}=W(\Enc(I)-\hat\mu)$,
with $W=V\diag(\lambda)^{-1/2}$ from the eigendecomposition of the encoded
covariance, is computed once on the training frames at the end of that stage and
frozen with the chart. Under multi-view supervision
(Section~\ref{sec:multiview}) the novel views are $65$ SHARP renders per frame,
computed once and cached. Stage~2 trains the dynamics alone on windows of the
clip, again for about $5\times10^4$ steps, its objective one differentiable
Verlet step with latent position and velocity weighted equally in the visibility
metric, ridge $\rho=0.01$. High-resolution renders retrain the
decoder alone on the frozen chart.

\bibliographystyle{plainnat}
\bibliography{lngs_refs}

\end{document}